\documentclass[sigconf]{acmart} %% final version

\AtBeginDocument{%
  }
    
\copyrightyear{2023} 
\acmYear{2023} 
\setcopyright{acmlicensed}\acmConference[CHI '23]{Proceedings of the 2023 CHI Conference on Human Factors in Computing Systems}{April 23--28, 2023}{Hamburg, Germany}
\acmBooktitle{Proceedings of the 2023 CHI Conference on Human Factors in Computing Systems (CHI '23), April 23--28, 2023, Hamburg, Germany}
\acmPrice{15.00}
\acmDOI{10.1145/3544548.3581320}
\acmISBN{978-1-4503-9421-5/23/04}

\usepackage{amsmath}
\usepackage{graphicx}
\usepackage{wrapfig}
\usepackage{subcaption}
\begin{document}

% \title{Ambiguous explanations: how humans perceive visual explanations}

\title{Graphical Perception of Saliency-based Model Explanations}

% graphical perception
% of saliency based model explanations
% explanation
% 


\author{Yayan Zhao}
\affiliation{%
  \institution{Vanderbilt University}
  \country{USA}}

\author{Mingwei Li}
\affiliation{%
  \institution{Vanderbilt University}
  \country{USA}}
  
\author{Matthew Berger}
\affiliation{%
  \institution{Vanderbilt University}
  \country{USA}}

\begin{abstract}
In recent years, considerable work has been devoted to explaining predictive, deep learning-based models, and in turn how to evaluate explanations.
An important class of evaluation methods are ones that are human-centered, which typically require the communication of explanations through visualizations.
And while visualization plays a critical role in perceiving and understanding model explanations, how visualization design impacts human perception of explanations remains poorly understood. 
In this work, we study the graphical perception of model explanations, specifically, saliency-based explanations for visual recognition models.
We propose an experimental design to investigate how human perception is influenced by visualization design, wherein we study the task of alignment assessment, or whether a saliency map aligns with an object in an image.
Our findings show that factors related to visualization design decisions, the type of alignment, and qualities of the saliency map all play important roles in how humans perceive saliency-based visual explanations.
\end{abstract}

% Matt's selected / generated concepts



\maketitle

\section{Introduction}

The widespread development and deployment of predictive models in recent years, particularly deep neural networks, has been met with an increasing demand for model transparency. Indeed, there is growing evidence that the comprehension of how models make decisions is a critical factor for their adoption in real-world settings \cite{shen2017deep,gawehn2016deep,goodman2017european,6907100}. As a concrete example, consider a machine learning model for visual recognition, where the goal of the model is to predict a category, given an image. Numerous methods have been developed to \emph{explain} such predictions, often realized by attributing importance to locations in the image that are most relevant to the model's reasoning~\cite{zhou2016learning,selvaraju2017grad}. These visual explanations play an especially important role in high-risk settings, e.g. within medical image processing, the identification of clinically-relevant features that explain a given prediction can serve as a valuable tool for clinical decision support~\cite{thakoor2022multimodal,dravid2022medxgan}.

Visual explanations have been deployed, and their effectiveness studied, across a diverse set of human-AI collaboration tasks~\cite{Fel2021WhatIC,10.1007/978-3-031-19775-8_17,nguyen2021effectiveness}. A commonality to these studies is the requirement of \emph{communicating} visual explanations to humans as part of solving a downstream task. There are numerous factors that underlie such communication: (1) the content of the explanation itself, (2) the visual encoding choices made by designers for displaying an explanation, and (3) how a human perceives the resulting visualization. The perception of an explanation is foundational to how a human builds an understanding of an explanation, and consequently, makes decisions in AI-assisted scenarios. Yet the role of perception has received comparatively less attention within the community, with most studies holding fixed the communication of explanations, and instead focusing on how well explanations enable humans to reason about model predictions~\cite{10.1007/978-3-031-19775-8_17,nguyen2021effectiveness}. We argue that a better understanding of perception can provide a more complete picture on model explanations, and can shed light on \emph{why} humans make judgements that concern machine learning models.

\begin{figure*}[!t]
    \centering
     \includegraphics[width=0.9\linewidth]{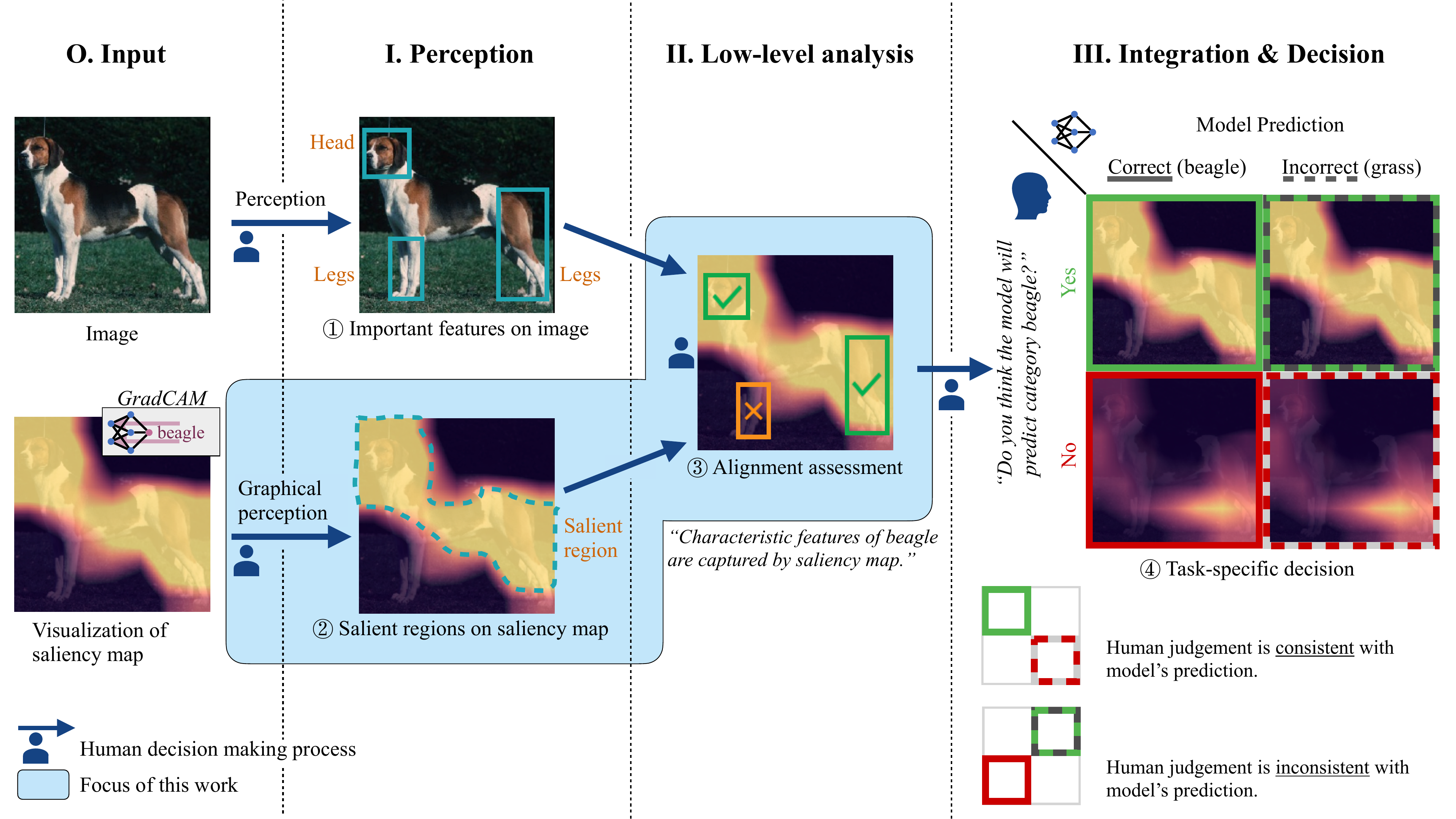}
     
    \caption{We place the focus of our work (highlighted in blue) within the broader context of human-AI collaboration. Given a particular task, e.g. decide on whether a model will predict a given category, (I.) a human will first perceive the image for task-relevant features on the input, and perceive the graphical encoding of the visual explanation for features important to the model. Next (II.) a human will assess alignment between image features, and model-derived features. From here (III.), a human will then make a decision on the provided task. Studying the perception of alignment provides us with a better understanding of human decisions made in AI-assisted tasks, whether a human's decision is consistent, or inconsistent, with a model's prediction.}
    \label{fig:overall-flow}
    \Description[A flow chart of human decision making process in human-AI collaboration]{Flowchart of how human make decision in the context of human-AI collaboration. It is composed of four phases: Input to human; Human perception of image and saliency map; Low-level analysis of human on the input; Integration and decision.}
\end{figure*}

In this paper, we focus on a core element of visualization design: \emph{graphical perception}~\cite{cleveland1984graphical}, placed in relation to visual encodings of model explanations. Our problem setting is the field of computer vision, wherein discriminative models are highly performant~\cite{7780459}, the development of model explanations is pervasive~\cite{zhou2016learning,selvaraju2017grad,petsiuk2018rise}, and the application of explanations, and in turn their evaluation, has received considerable attention~\cite{Fel2021WhatIC,10.1007/978-3-031-19775-8_17,nguyen2021effectiveness}. We study \emph{saliency map} explanations, methods that assign a value of importance to each pixel in the image, where high values suggest locations that are important to the model for prediction. We investigate how characteristics of the saliency map, as well as visualization design choices, influence human judgement on the task of \emph{alignment assessment}, namely, whether the saliency map faithfully captures a given object within an image.

We view perception of alignment as playing an important, although not complete, role in how humans engage with model explanations. Fig.\ref{fig:overall-flow} contextualizes the focus of our work within the broader scope of human-AI collaboration. Building on prior studies~\cite{fel2022harmonizing,ullman2016atoms,dicarlo2012does}, for a given AI-assisted task, we decompose the activities that a human performs into three phrases. First (Fig.\ref{fig:overall-flow} I.), a human will perceive the image to identify important features (e.g. head and legs of a beagle), in addition to perceiving the visually-encoded saliency in identifying regions deemed important by the model (e.g. head, body, and hind legs). Secondly (Fig.\ref{fig:overall-flow} II.), a human will assess the alignment between features extracted from the image, and features extracted from the saliency map, e.g. the saliency map captures the majority of beagle-relevant features. Last (Fig.\ref{fig:overall-flow} III.), from their understanding of the explanation, a human will make a decision on model behavior, e.g. whether they think the model will predict the category beagle. In some cases, a human's judgement will be consistent with the model's prediction, e.g. what a human thinks will be predicted by the model is in fact the model's prediction. In other cases, inconsistency between human judgement and model prediction can arise, e.g. a human decides the model will not predict beagle, when in fact the model predicts beagle. To better discern situations where humans and models are either consistent or inconsistent in their decisions, it is critical to understand how humans perceive explanations, and in particular, the graphical perception of their visual displays. For instance, in Fig.\ref{fig:overall-flow} III. the model might predict the category grass, but due to an acceptable alignment, a human would decide the model predicts beagle (top-right cell); conversely, the model predicts category beagle, yet due to misalignment, a human would decide the model's prediction is not beagle (bottom-left cell).

To better understand the matter of alignment assessment, our experimental design investigates what factors contribute to the graphical perception of saliency-based explanations. We hypothesize that choices made in visualization design can have a significant impact on how humans perceive alignment. Specifically, we study visual encodings of saliency maps along two axes: the choice of the encoding's \emph{visual range}, and choice of its \emph{data domain}. We seek an understanding of how human judgement varies based on the \emph{amount of detail} that a visual encoding depicts, both in its range -- namely, a binary mask, a contour-based visualization, and a heatmap, as well as in its domain -- selecting an interval of saliency values to be displayed. Beyond choice of visualization, our experimental design is further organized around characterizations of the saliency map, the shape of the object present in an image, and the type of alignment between the saliency map and the object. We conduct a user study to elicit human judgement on alignment assessment, and in turn, identify factors most influential to graphical perception. Moreover, we build perceptual models that serve as proxies for human judgement in perceiving saliency maps.

The findings, and contributions, from our user study are summarized below:

\begin{enumerate}
\item We find a significant effect of visual encoding on human judgement, suggesting that quantitative information afforded by heatmaps and contours is used by humans, in contrast with binary masks. 
\item We find that the type of alignment is important, e.g. humans will agree with saliency maps that overestimate objects more so than underestimate. 
\item We find the distribution of values in the saliency map influences human judgement -- saliency maps that are ``binary-like'' are more likely to elicit positive responses of alignment.
\item Last, we contribute a perceptual model of human judgement, tailored to the type of visualization and choice of its parameters. The model permits us to compare parameter settings of visual encodings, wherein we find a clear delineation across different types of alignment.
% \item Last, we illustrate how our perceptual model of human judgement can inform visualization design, accomplished by providing ``scents''~\cite{willett2007scented} on parameter settings that anticipate how likely humans would positively respond to particular visualization parameters.
\end{enumerate}

\section{Related Work}

We discuss work related to model explanations, their evaluation, and perceptual factors that underlie the visualization of quantitative data.

\subsection{Model explanations}

Methods that help humans understand machine learning models have witnessed significant progress in recent years, motivated by their application in high-risk areas~\cite{6907100,shen2017deep} where trust in models is important. 
In particular, for visual recognition models, significant work has been devoted to interpreting their learned representations~\cite{zeiler2014visualizing,DBLP:conf/cvpr/DosovitskiyB16}, as well as probing their representations for specific forms of knowledge~\cite{DBLP:conf/cvpr/BauZKO017,8081219,fong2018net2vec,10.5555/3495724.3497163}. Most relevant to our work are methods that attempt to explain the predictions of visual recognition models. 
In this paper, we focus on saliency methods, also known as feature attribution methods, which assign importance to input features. There are mainly two branches of saliency methods: perturbation methods and backpropagation methods. Perturbation methods~\cite{zeiler2014visualizing,ribeiro-etal-2016-trust, kapishnikov2019xrai,fong2017interpretable, 9010039,dabkowski2017real} compute the importance of each pixel on the image by measuring the model's response when perturbing the image. For example, RISE~\cite{petsiuk2018rise} probes a model with randomly masked versions of the input images, and generates saliency maps based on the sum of these random masks weighted by their corresponding prediction scores. The method of LIME~\cite{ribeiro-etal-2016-trust} learns
linear weights corresponding to the effect of perturbations on different image patches in an image.
Backpropagation methods~\cite{simonyan2013deep,smilkov2017smoothgrad,springenberg2014striving} compute the importance of each pixel by computing gradients, or modifications of gradients, of designed functions that aim to represent model behavior.
Among them, the family of methods known as class activation mapping (CAM)~\cite{zhou2016learning,DBLP:conf/iclr/AnconaCO018,selvaraju2017grad} attributes importance based on the gradient of model predictions. These methods produce images that represent a notion of \emph{saliency}, locations in the image most relevant to the model's prediction.

In particular, we study how humans perceive such saliency maps, limited to a particular explanation method: GradCAM~\cite{selvaraju2017grad}. This method produces a coarse, low-resolution localization map of important regions to the model's prediction. We view GradCAM as most representative of the CAM family~\cite{zhou2016learning}, all methods of which similarly produce low-resolution saliency maps~\cite{10.1007/978-3-030-58555-6_37,DBLP:conf/iccv/KimCAO21,9878440,cole2022label} that aim to explain model predictions. These methods often evaluate the quality of explanations in terms of their methodology, and ability to localize objects in images, but typically do not verify if such evaluation metrics are commensurate with human perception.

\subsection{Evaluation of visual explanations}

There has been a diverse set of approaches considered for evaluating model explanations, which can roughly be categorized into two groups, as discussed more broadly in Doshi et al.~\cite{doshi2017towards}. One line of work conducts evaluation by assuming known, ground-truth explanations of model predictions exist, e.g. binary masks of objects relevant to a category, and develops metrics for scoring the localization quality of saliency maps, relative to object masks~\cite{zhou2016learning,selvaraju2017grad,zhang2018top}. To derive a metric of localization, it is often common to first post-process saliency maps to yield binary masks~\cite{DBLP:conf/cvpr/BauZKO017,fong2018net2vec,10.5555/3495724.3497163} and second to score the quality of extracted binary masks against ground-truth binary object masks~\cite{DBLP:conf/cvpr/BauZKO017,DBLP:conf/iclr/BauZSZTFT19,10.1145/3491102.3501965}. Further, perturbation-based modifications have been considered~\cite{adebayo2018sanity,heo2019fooling} to improve robustness of automatic evaluation metrics.

The process of (1) utilizing a collection of object masks as ground-truth explanations, (2) deriving binary masks from saliency maps, and (3) evaluating the quality of these masks through automatic evaluation scores, is intended to serve as a proxy for human evaluation. However, a separate line of work has demonstrated that human-centered evaluation of model explanations might not always agree with automatic evaluation methods. These experimental designs are usually task-oriented, e.g. 
% assessing whether visual explanations are indicative of model predictions~\cite{nguyen2021effectiveness}, 
understanding failure modes~\cite{Fel2021WhatIC}, the ability to simulate model behavior~\cite{10.1145/3377325.3377519}, or assisting users in improving their own reasoning for recognition tasks~\cite{Shen2020HowUA,Alufaisan_Marusich_Bakdash_Zhou_Kantarcioglu_2021}.
For example, HIVE~\cite{10.1007/978-3-031-19775-8_17} provides users with an input image and a collection of model explanations, and asks users to identify the model's output, and select the correct prediction, amongst the explanations. The ground-truth class label is concealed to prevent users from relying on prior knowledge, and instead, used to evaluate explanations based on human responses. Nguyen et al. ~\cite{nguyen2021effectiveness} provides users with an input image and the prediction of the model for that image, along with the heatmap visualization of a saliency map explanation. Users are asked to decide whether the model prediction is correct, wherein explanations are evaluated based on user performance. More broadly, existing work has studied the consequences of poor model explanations, ranging from loss of agency~\cite{levy2021assessing}, to misplaced trust~\cite{nourani2019effects,poursabzi2021manipulating}. Our work aims to complement such existing human-centered evaluation methodologies by studying, what we argue, is a precursor to reasoning about a presented task: the perception of visual explanations, and specifically, how these explanations are visualized.

\subsection{Perceptual factors in visualizing quantitative data}

Within the visualization community significant work has been conducted in studying graphical perception of quantitatively-encoded data. Often such experiments are distinguished by the task presented to humans, ranging from ratio assessment~\cite{cleveland1984graphical}, judgement of correlation~\cite{harrison2014ranking}, to estimation of averages amongst a set of graphical marks~\cite{yuan2019perceptual}. Our experimental setup is reminiscent of lineup-based experimental designs~\cite{wickham2010graphical,beecham2016map}, wherein stimuli are arranged in a small-multiples view, and users must perform comparative judgements. Moreover, significant work has studied the design~\cite{bujack2017good} and human perception for color-mapped scalar fields, assessing factors such as spatial frequency~\cite{10.1145/3173574.3173846}, quantization~\cite{padilla2016evaluating}, and color map designs~\cite{dasgupta2018effect} for various tasks, e.g. value judgements and comparison-based judgements~\cite{10.1145/3173574.3173846}. Our work, similarly, studies visualization design of 2D scalar fields, using color maps in addition to the display of isocontours, but our task differs, in that we study how visualization effects perception of saliency in the localization of a known object in the image.

\section{Perception of Alignment: Task and Hypotheses}
\label{sec:hypotheses}

\subsection{Task}\label{sect:task}

The overarching goal of our work is to study the effect that visualization design has on the perception of visual explanations, namely, saliency maps. Towards this end, it is necessary to present users with a task that is, at once, pertinent to the evaluation of explanations~\cite{10.1007/978-3-031-19775-8_17}, while mitigating confounding factors that are not relevant to visual perception, but nevertheless influence human judgement, e.g. expectations of model behavior. Given these considerations, we employ the following task in our study: \textbf{is a given saliency map well-aligned with an object present in an image?} We clarify the notion of ``alignment'' with the following criteria:
\begin{enumerate}
    \item Do salient regions in the visual explanation, in large part, cover the presented object?
    \item Are these salient regions, in large part, limited to the object?
    \item Do salient regions cover the most characteristic portions of the object?
\end{enumerate}
The first two criteria require humans to first identify important regions in the saliency map, and then assess the overlap of these regions with an object, and \emph{only} the object. The last criterion is intended to help humans make judgements on partially-aligned saliency maps, giving priority to those that capture the most relevant properties of an object. We expect this to cause small inter-participant variance due to subjectivity on what constitutes a characteristic part of an object.

\begin{figure}[!t]
\centering
    \includegraphics[width=0.95\linewidth]{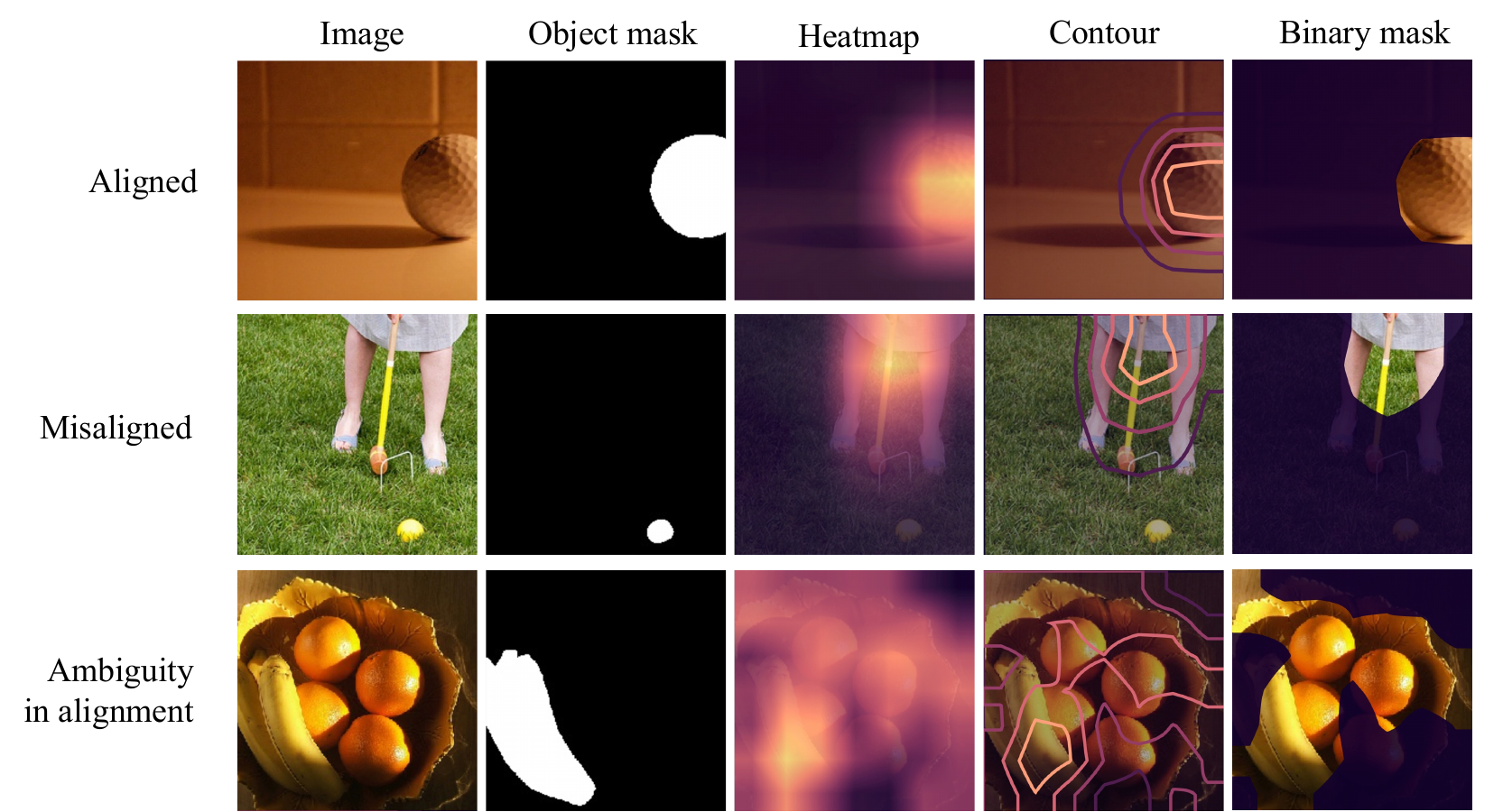}
    \caption{In cases where saliency-based model explanations align well with the object (top row) or not (middle row), we do not anticipate visualization playing a role in perception. However, when model explanations are imperfect, giving rise to ambiguity in alignment (bottom row), we anticipate that \emph{how saliency maps are visually encoded} will impact human perception.}
    \label{fig:alignment}
    \Description[Examples of images and different visualizations of their saliency maps, for different alignment types between images and saliency maps.]{A figure showing examples for three different alignment types between objects on the image and saliency maps: aligned, misaligned, and ambiguity in alignment. For each example, we show the image, the binary mask of the object on image, the heatmap visualization of saliency map, the contour visualization of saliency map, and the binary mask visualization of saliency map.}
\end{figure}

\textbf{Motivation:} human judgement of alignment might not seem to be a task that would heavily depend on visualization design. Indeed, numerous metrics exist that compute the alignment between activations of a network, and the ground-truth mask of a set of concepts \cite{DBLP:conf/cvpr/BauZKO017,fong2018net2vec,10.5555/3495724.3497163}, or the alignment between a saliency map and ground-truth segmentation mask for a given image \cite{selvaraju2017grad}. It is common for such methods to (1) extract a binary mask via thresholding a continuously-valued visual explanation, and (2) measure the alignment of this mask with the ground-truth object mask. When there is little ambiguity on whether or not a visual explanation is aligned with an object, then such metrics are likely to be good proxies for human judgement (c.f. Fig.~\ref{fig:alignment}), regardless of the choice of visual encoding. However, when explanations present ambiguity for humans in assessing alignment, then we expect variability across visualization designs (c.f. Fig.~\ref{fig:alignment}), and thus, a metric computed \emph{irrespective} of visualization may no longer accurately represent human judgement.

\subsection{Hypotheses}

Our experimental design is driven by the following hypotheses that we seek to verify through our study:

\textbf{\underline{(H1)} The type of visual encoding will influence human perception of visual explanations.} We distinguish visual encoding types by the level of detail that can be presented in their visual range
(c.f. rows in Fig~\ref{fig:amount-of-info-and-parameters})
. Specifically, a \textbf{binary mask} corresponds to the smallest amount of detail shown, namely, selectively displaying pixels whose saliency values exceed a given threshold. A \textbf{contouring} of the saliency map shows a discrete number of saliency values, each encoded as curves of constant saliency. Last, a \textbf{heatmap} visually encodes each saliency value in the image by a continuous color scale. We anticipate that human judgement will be impacted by the level of detail (two values, discretization, continuous) offered by an encoding's visual range.

\textbf{\underline{(H2)} The type of alignment will impact perception in a data-dependent manner.} Motivated by Boggust et al.~\cite{10.1145/3491102.3501965}, we distinguish between two types of alignment: saliency maps that \emph{overestimate} an object, and saliency maps that \emph{underestimate} an object. We anticipate that the \textbf{data range} chosen as the domain for a visual encoding will influence perception on alignment type
(c.f. columns in Fig~\ref{fig:amount-of-info-and-parameters})
, in order to selectively narrow the range of data shown, or widen the range of data shown.

\textbf{\underline{(H3)} The qualities of the saliency map will influence perception of visual encodings.} Viewed as a sampled 2D scalar field, the saliency map can be characterized in terms of its \emph{distribution of values} \cite{sibrel2020relation}. For instance, a saliency map comprised of just 2 values (0/1), will lead to different visual patterns compared to a saliency map whose values are uniformly distributed. Further, under a fixed distribution of values, we hypothesize that visual patterns will be perceived differently according to the type of visual encoding.

\textbf{\underline{(H4)} The shape of the object in an image will impact perception of alignment.} We hypothesize that the overall \emph{complexity} of the object shape will influence human judgement of saliency map alignment  \cite{tuch2009visual}. For instance, simple shapes, e.g. those that are convex, are easier for making judgements of alignment, in comparison to shapes that are highly nonconvex.

\section{Experimental Design: Methods}
\label{sect:methods}

To study human perception of saliency maps, we consider the following plausible factors that may influence perception:
1) the visual encoding of the saliency map; 
2) the characteristics of alignment between a saliency map and the given object; 
3) the level of detail present in saliency maps and;
4) The shape of the object in the image.

\subsection{Visually encoding saliency maps}\label{sect:parameters}

\begin{figure}[!t]
    \includegraphics[width=\linewidth]{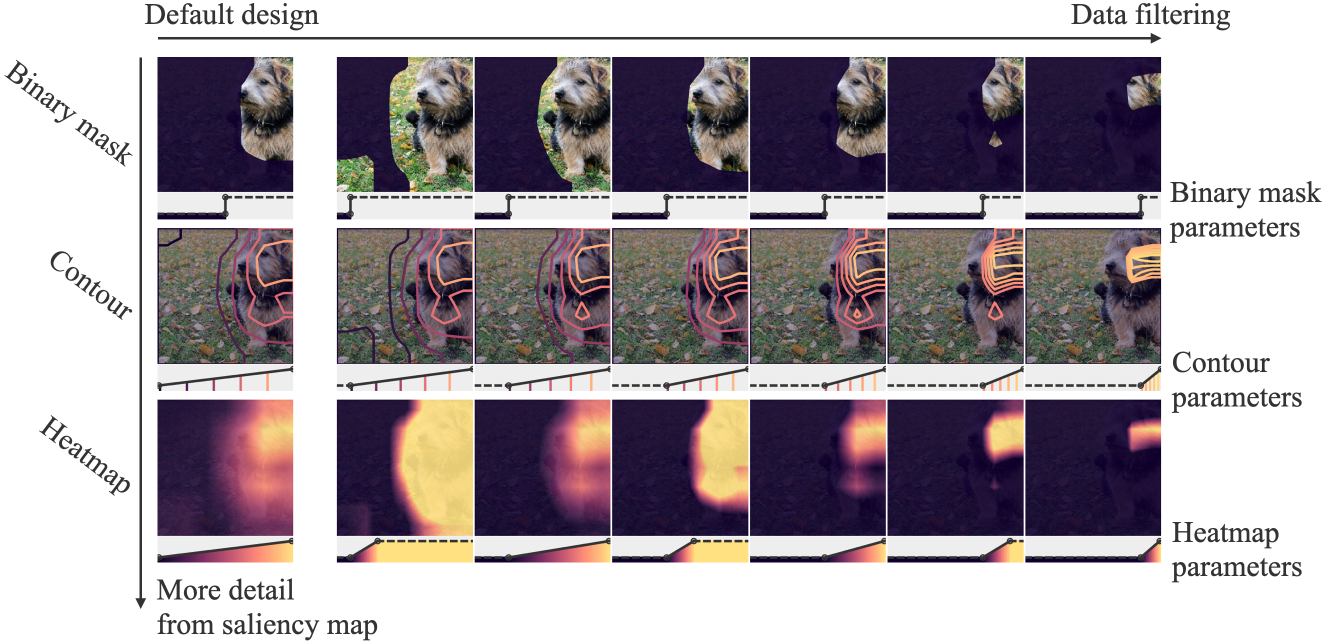}
    \caption{We study visualization designs along two axes: different rows indicate the amount of detail from a saliency map that a visualization can display, here shown in increasing amount through binary masks, contour-based visualizations, and heatmap visualizations. Different columns convey the amount of detail specified for the data domain in a visualization.}
    \label{fig:amount-of-info-and-parameters}
    \Description[A figure showing visualization designs of saliency maps along two axes: data filtering and visualization types.]{A figure showing the visualization designs for three different visualization types: heatmap, contour, and binary mask. For each visualization type, we show how the visualization of the same saliency map is changed along the axis of data filtering.}
\end{figure}

We wish to study the visual encoding of saliency maps along two distinct axes: (1) the level of detail afforded by the \emph{visual range} of a visual encoding, and (2) the level of detail that may be specified in the \emph{data domain} of a visual encoding, please see Fig.\ref{fig:amount-of-info-and-parameters} for an overview.

In increasing level of detail, we have chosen to visually encode saliency maps through: binary masks derived from saliency value thresholds, sets of curves produced through iso-contouring the saliency map, and heatmaps that densely, and continuously, encode saliency values. Our choice of binary masks is driven by their prevalence in automatic evaluation~\cite{DBLP:conf/cvpr/BauZKO017,DBLP:conf/iccv/KimCAO21}, as well as in human-centered explanation~\cite{10.1145/3491102.3501965}. Binary masks permit the \emph{smallest} amount of usable detail in saliency maps that can be displayed in a visualization -- regions in the image that have higher, or lower, saliency values. A contour visualization allow us to control for a \emph{discrete} amount of saliency detail to be displayed, namely, a small set of curves that each encode distinct saliency values. Contours are used in visualization design to summarize fields, emphasize geometric features, and non-obtrusively overlay graphical marks on top of other views~\cite{mayorga2013splatterplots,sarikaya2017scatterplots}. A heatmap permits the widest amount of displayable detail from saliency maps, e.g. every location in the image is assigned its own saliency-encoded color. Although heatmaps are, arguably, the predominant visualization for saliency evaluations~\cite{selvaraju2017grad,adebayo2018sanity}, prior work has studied how the precision at which heatmaps are encoded might have little effect on perceptual tasks~\cite{padilla2016evaluating}, where we view our contour-based encoding as an alternative to quantizing heatmaps.

Our method for specifying data domain, and consequently controlling the amount of data to be encoded, varies by visual encoding. For a binary mask, the value at which to threshold the saliency map is treated as a data filtering parameter which belongs to one of the visualization parameters we choose.
A given threshold will result in selectively showing only those regions in the image whose saliency values exceed the threshold. Within a contour-based visualization, we allow for specifying a single saliency value, representing the minimum saliency at which to display contours. An equal-spaced sequence of saliency values are then generated, ranging from the prescribed value to the largest value in the saliency map, and for each, an iso-contouring of the saliency map is produced. Last, for a heatmap we allow for the specification of two parameters -- a minimum saliency value and maximum saliency value -- which determines the data domain for the color scale. We use a luminance-increasing color scale designed in LAB color space, where dark colors indicate low saliency and bright colors indicate high saliency. The color scale further sweeps through hues -- from red to yellow -- in increasing saliency value, analogous to a black body color scale~\cite{moreland2016we}. In addition, we use this same scale to map color onto saliency-specific curves within the contour visualization.

Although these different visualization designs provide unique ways to visually encode data (i.e., masked regions, curves and continuous colors) under varying specifications of data (single value, discrete levels, continuous interval),
the notion of a \emph{transfer function} allows us to view their color mappings in a unified way, as shown in the line plots below each visualization design in Fig.\ref{fig:amount-of-info-and-parameters}.
Formally, we assume the saliency values $t$ in a saliency map to be normalized to $[0,1]$,
and consider a piece-wise linear function that maps saliency value $t \leq p_0$ to $0$, $t > p_1$ to $1$ and linearly interpolate in between.
The range of the transfer function is further mapped to the fixed color scheme and we use this as the basis for color mappings in all three visualization types. 
In binary mask, we control the threshold by letting $threshold = p_0 = p_1$, (c.f. top row, Fig.\ref{fig:amount-of-info-and-parameters}).
For the contour visualization, we fix the color mapping by letting $p_0 = 0$, $p_1 = 1$. 
We also fix the number of levels shown in contour plot to be $N = 5$, to encode a moderate level of detail shown in the contour visualization. 
Then, we control the base level of the contour, $\text{b}$, and show equally spaced levels in the contour plot from the base level to the maximum level (c.f. middle row, Fig.\ref{fig:amount-of-info-and-parameters}).
In the heatmap visualization, clamping maximum values in the data domain can be beneficial, in displaying a wider range of colors for a narrower range of data values.
We control the saliency value range $(p_0, p_1)$ that will be color-encoded in the visualization (bottom row, Fig~\ref{fig:amount-of-info-and-parameters}).
In a given saliency map, these two parameters control the range of saliency values beyond which the heatmap saturates, and one may see them as ways of removing outlying saliency values from the data.
Another interpretable re-parameterization of $(p_0, p_1)$ is $(\frac{p_0+p_1}{2}, p_1 - p_0)$, where $\frac{p_0+p_1}{2}$ is the saliency value that is encoded into the middle color in the color map, and $p_1 - p_0$ represents the extent of the encoded saliency values in the data domain.

\subsection{Computing and categorizing alignment}\label{sect:categorize-alignment}

Beyond visualization design, other criteria that we believe impacts human perception of alignment is (1) the \emph{level} of alignment between a saliency map and a given object, and (2) the \emph{type} of alignment. However, the computation of such factors \emph{independent} of humans might not accurately reflect their judgement. To this end, we approach the \emph{level} of alignment as a continuous value, in order to \emph{factor out} alignment level in our study, while we approach \emph{type} of alignment through a conservative grouping criteria based on prior metrics.

Specifically, we build on Boggust et al.~\cite{10.1145/3491102.3501965}, wherein for a given image with a segmentation mask of a ground truth object $G$, and a binarized saliency map $S$ obtained through appropriate thresholding, 
we compute three alignment measures: ground-truth coverage (GTC), saliency coverage (SC) and intersection-over-union (IoU),
where each metric is in the range $[0,1]$.
Intuitively, high ground-truth coverage (GTC) measure implies that the saliency map, in large part, covers the ground truth feature, but may overestimate the ground truth;
a high saliency coverage (SC) implies the saliency map, in large part, sits within the ground truth object mask, but may underestimate the ground truth;
and when both GTC and SC are medium, the saliency map partially overlaps with the ground truth.
Based on these intuitions, in this study we select and group instances of saliency maps into three alignment types: overestimation, underestimation and partial alignment.
We hypothesize that these types of alignment will influence human perception of saliency maps.
Moreover, the intersection-over-union (IoU) will only return a high score when, \emph{both}, the thresholded saliency map is contained within the object, and limited to the object -- we use IoU as our level of alignment.

\begin{figure}[!t]
    \includegraphics[trim={0 0 3.6cm 0},clip,width=0.85\linewidth]{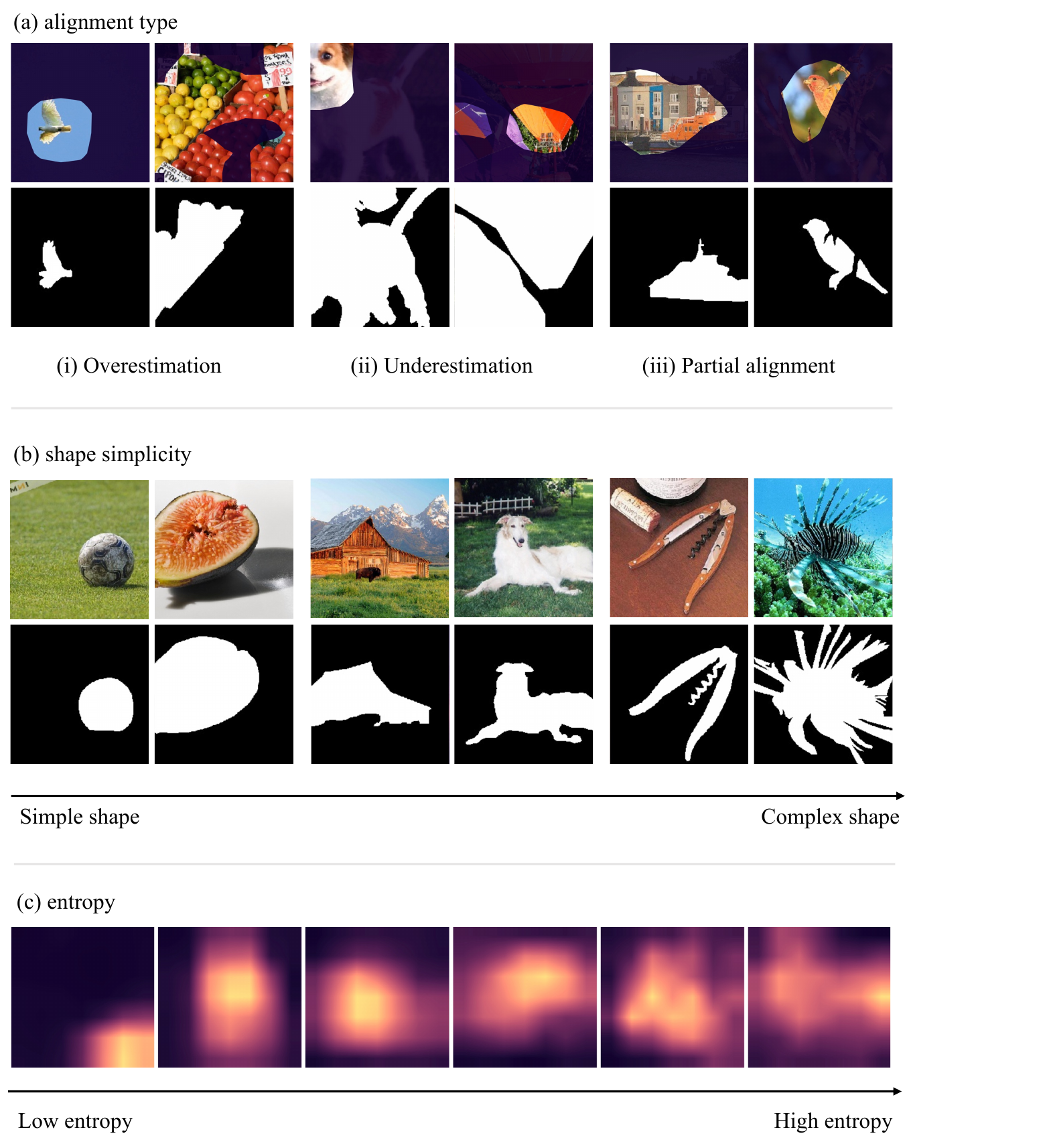}
    \caption{ a) \textbf{Alignment type}: We categorize alignment into three types: (i) overestimation; (ii) underestimation and (iii) partial alignment. b) \textbf{Shape simplicity: }We measure shape complexity by ratio of two perimeters. Left to right: simple shapes to complex shapes according to the shape simplicity measure. c) \textbf{Entropy}: We characterize saliency maps by the spatial distribution of saliency values through entropy. Left to right: low entropy to high entropy, shown as heatmaps.}
    \label{fig:entropy-alignment-shape}
    \Description[Examples for different alignment types, shape complexity, and entropy of saliency maps.]{A figure showing a) examples of different alignment types between objects on images and saliency maps; b) examples of objects on images with shape from simple to complex; c) examples of saliency map with entropy from low to high.}
\end{figure}

%% how and why we threshold
Computing GTC, SC, and IoU requires a choice of threshold for each saliency map.
The choice can be critical for the numerical values of these measures, but may not have as much influence on the relative ranking of instances and even less so on a coarse grouping of alignment types.
Boggust et al.~\cite{10.1145/3491102.3501965} uses one standard deviation above the mean saliency value as a threshold. 
In our study we use a threshold related to the mean saliency value, but more connected to the standard way of visualizing saliency maps in a heatmap,
where we take the mean of the minimum and maximum saliency value in a saliency map as the threshold.

% How we group alignment types
% 1. Choose IoU range 0.05 - 0.68, where alignment is ambiguous and may depend on visualization
% 2. Define:
% Over: sc(0 - 0.25), gtc(0.75 - 1) ~5738 samples;
% Under: sc(0.75 - 1) gtc(0 - 0.25) ~1079 samples;
% Partial: sc(0.35 - 0.65) gtc (0.35 - 0.65) ~9940 samples;
To group instances into three alignment types, we first group saliency maps (along with their associated ground truth mask) into three disjoint intervals along the GTC and SC measure.
Specifically, we group instances by GTC or SC into low (0 - 0.25), medium (0.35 - 0.65) and high (0.75 - 1) levels. 
Then, we categorize a saliency map as \emph{overestimating} the ground truth if it has high GTC and low SC; similarly we define \emph{underestimation} as low GTC and high SC; we define \emph{partial alignment} as medium GTC and SC.
The intuition is that when the ground truth is well covered (high GTC) by saliency but the saliency is only partially covered by ground truth (low SC), the saliency map overestimates the ground truth; a similar intuition holds for other types.
Furthermore, when sampling overestimated or underestimated examples, we exclude those that have extremely low IoU (in our case, IoU < 0.05), as they are unlikely to present ambiguity when displayed by any visualization.

\subsection{Characterizing saliency maps}

Many works from psychology and geographic information systems have found that the spatial structure of a continuous map will influence human understanding of heatmap-based visualizations of 2D field-based data.
For example, Sibrel et al.~\cite{sibrel2020relation} found that fields with a weak spatial structure, e.g. where values are uniformly spatially distributed, result in perceptual biases that differ from more distinct spatial structures, e.g. sharply-peaked regions in the image.
Since these varying spatial structures are also commonly seen in saliency maps, we hypothesize that the spatial structure of saliency maps would influence human understanding of saliency maps. 
Therefore, we consider a metric that describes the spatial distribution of a saliency map and use it as a factor in our study.

To derive a metric that characterize the spatial distribution, we measure the variation of saliency levels within a saliency map by its information entropy. 
In practice, we find entropy correlates well with high concentrations of saliency values within a saliency map. 
When visualized in heatmaps with a fixed color scale, low-entropy yields highly concentrated saliency regions, shown in high contrast (Fig. \ref{fig:entropy-alignment-shape}).

\subsection{Measuring shape complexity}

Works from psychology~\cite{tuch2009visual} have observed that objects with complex shapes will take more time to process, take more cognitive effort, and can be hard to remember. 
Based on these findings, we hypothesize that simple convex shapes are easier to process and compare against each other whereas highly non-convex shapes are more challenging to process.

There exists a number of metrics that measure 
shape complexity, such as the number of independent turns, angular variability \cite{attneave1957physical}, the perimeter of shape \cite{psarra2001describing}, the irregularity of shape \cite{berlyne1958influence}, and the variance of side lengths and largest radial length. 
In this work, we measure shape complexity by the ratio of two perimeters: the perimeter of the object outline, and the perimeter of its convex hull.
Intuitively, this measures the complexity of the shape against a convex counterpart, and we bound the range of the measure by using convex-hull-to-object-outline ratio instead of the inverse.
Therefore, the resulting measure ranges from $[0,1]$: the larger the measure, the simpler the shape.

\section{User Study}

Here we describe the data preparation, interface, and experimental conditions used as part of our study.

\subsection{Preparation} 

\subsubsection{Data: Images and Masks}
We select images from the ImageNet Large Scale Visual Recognition Challenge (ILSVRC)
 \cite{5206848}, a popular large scale dataset with each image assigned a label from 1 of 1,000 categories.
For our purposes, however, ImageNet alone is insufficient, as objects present within images are only captured by bounding boxes, a rather coarse annotation for the analysis of shape complexity.
To faithfully capture the shape of the object, we adapt the ImageNet-S dataset \cite{gao2021large}, a subset derived from ImageNet instances that have been annotated with object segmentation masks.
From the validation set of ImageNet-S which includes 12,419 images in 919 categories, we exclude images that we believe would be unfamiliar, offensive, or otherwise inappropriate to display to participants, based on their category.
From the remaining subset, we further exclude those that are low resolution, or contain hard-to-detect objects, e.g. animals camouflaged within a scene.
Finally, we augment the dataset by random scaling and cropping, and remove low quality crops from the augmented dataset, which includes either tiny, oversized or overly trimmed objects in those particular crops.

\subsubsection{Saliency Maps}
We derive saliency maps by applying GradCAM~\cite{selvaraju2017grad} to a pre-trained ResNet-50 model~\cite{7780459}.
For each image instance, we compute GradCAM from its top-5 predicted labels, as the top-5 predictions are commonly used in measuring model performance.
If an image is mis-classified by the model, we also include the GradCAM computed from the ground truth label, resulting in 5--6 saliency maps per image.
Next, we select unique-looking saliency maps based on the graph of similarities between saliency maps.
Specifically, for every set of saliency maps of a given image, we measure pairwise Euclidean distance between the normalized saliency maps. 
We consider two saliency maps similar if their distance is below a small threshold and connect them by a link.
Among saliency maps that are path-connected by such links, we randomly select one as the representative.
Together with saliency maps that are dissimilar from others, we derive on average 3 unique saliency maps per image.

\begin{figure}[!t]
    \includegraphics[width=\linewidth]{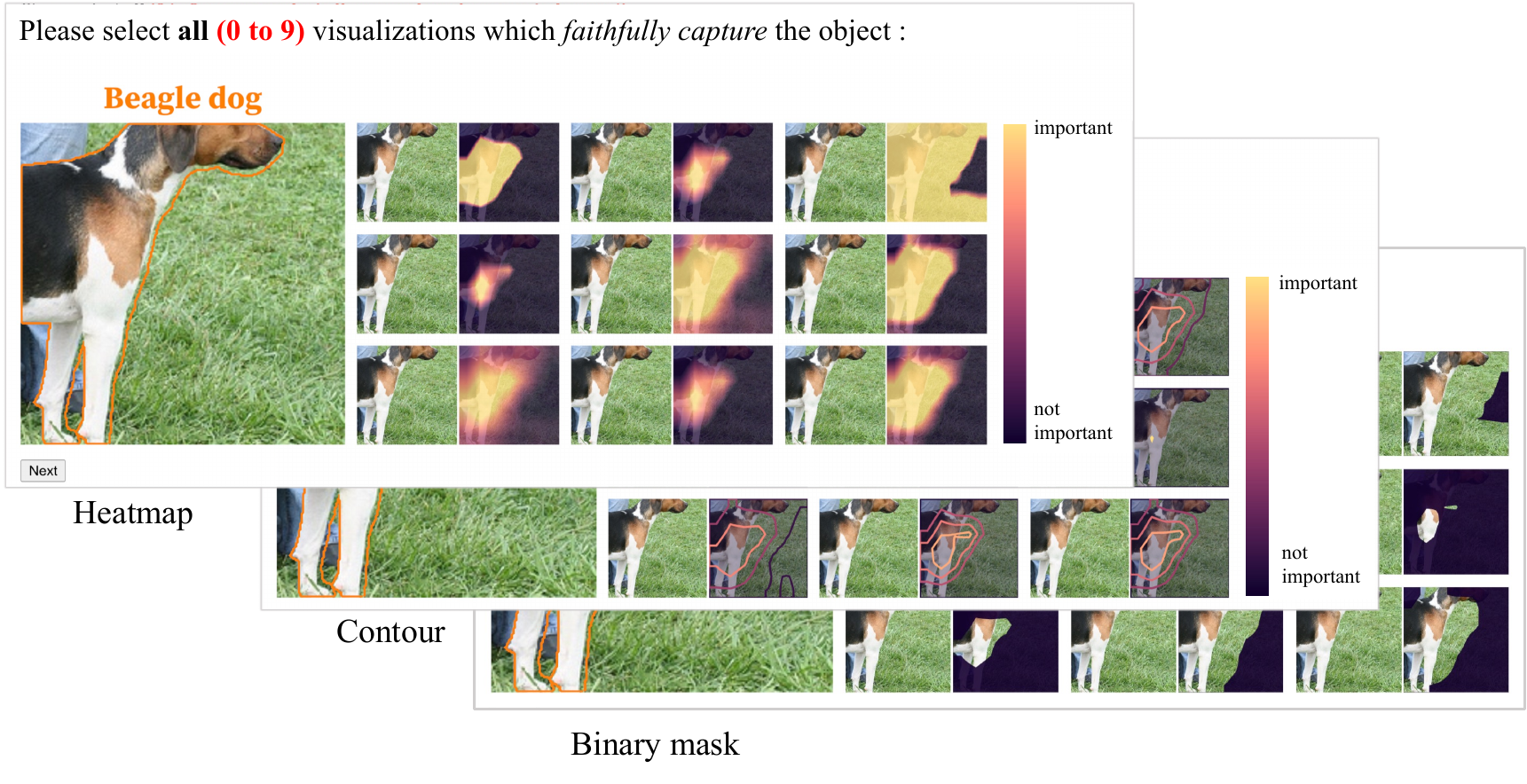}
    \caption{User interface in the study.}
    \label{fig:ui}
    \Description[User study interface]{Screenshots of user interfaces for three different visualization types in the user study.}
\end{figure}

\subsubsection{Interface}
% Explain the UI:

A central goal in the design of our interface is to ensure participants can make well-informed decisions on whether or not a given saliency map is aligned with an object in an image. Early experiments suggested that showing one visualization at a time leads to high variance in responses, and a strong dependency on the order in which saliency maps are presented to participants. Hence, to provide context, in our interface a single trial consists of showing a participant a single image, visualized under 9 different parameter settings drawn from the visualization's data domain (c.f. Sec.~\ref{sect:parameters}). We arrange the visualizations within a $3 \times 3$ grid, positioned at random grid cells in order to inhibit user dependencies on specific grid cells. Each grid cell juxtaposes the original image with the saliency map visualization. To help participants quickly recognize the object in question, we show a single higher-resolution image, with a curve encompassing the object superimposed on the image (c.f. Fig.\ref{fig:ui}). For a single trial, a participant is asked to select \emph{any} visualizations (potentially none) that, in their best judgement, faithfully capture the object present within the image (c.f. Sec.~\ref{sect:task}). We record the visualizations chosen by the user, along with the time taken to complete a trial.

\paragraph{Discussion:} our interface design bears similarity to the lineup-based interfaces employed in graphical perception studies~\cite{wickham2010graphical}. However, where these methods test the ability of people to make statistical assessments provided a small multiples visualization, our interface intends to provide context for users to make \emph{relative judgements} within a visualization's design space (choice of data domain). This provides us with signal on \emph{how much} of the design space a user agrees with -- number of visualizations chosen -- along with the specific data domains that underlie these visualizations. One potential design alternative is to find the parameter setting at which users are unable to judge whether one visualization better aligns with an object, compared to another visualization, akin to just-noticeable differences (JND). In this scenario, pairwise comparison judgements are suitable, along with an adaptive means of sampling visualizations to quickly arrive at such a judgement. Staircasing techniques used to estimate JNDs are highly relevant for this task~\cite{harrison2014ranking,beecham2016map}. However, in our scenario we do not have a meaningful baseline from which to sample, e.g. correlation measures as used in~\cite{harrison2014ranking,beecham2016map}, and thus adaptive sampling becomes nontrivial to perform.

\subsubsection{Visualization design parameters}\label{sect:parameters-sample}

As discussed, in each task we sample 9 visualization design parameters for the same visualization type.
Recall from Sec.~\ref{sect:parameters} that the parameters are: threshold for the binary mask, minimum saliency level for the contour plot, and two clamping points (minimum and maximum saliency) for the heatmap.
Since a participant only observes a small sample of 9 designs for each task, and we want the task to be well-contextualized within the design space, we apply stratified sampling over 9 equally-sized regions in the design space.
For binary mask and contours, this only requires one design parameter per plot.
We opt to perform stratified sampling to draw parameters: we uniformly bin the data domain $[0,1]$ into 9 intervals, and for each bin, we draw a random value from its interval as the parameter for that bin.
%For these two types, the design space is the $[0,1]$ interval that corresponds to the (normalized) saliency value domain.
%Therefore, we apply stratified sampling over 9 equal-length intervals, $\{\;(\frac{i}{9},\frac{i+1}{9}) \text{ for } i=0,1,\dots 8 \;\}$ in the design space. 
The heatmap visualization requires a pair of clamping parameters $(p_0, p_1)$ in the data domain. 
Recall that without loss of generality, we let $p_0 \leq p_1$. 
% This leads to a triangular design space (c.f. Fig. ~\ref{fig:voronoi-sample}). MSB: defer to supplemental
Given the triangular-shaped area, to sample 9 points in this 2D space, we perform stratified sampling over a Voronoi diagram comprised of 9 cells (c.f. Fig.\ref{fig:model2-alignment}).
Specifically, we use a constrained centroidal Voronoi tesselation, following~\cite{balzer2009capacity}, to partition the space into 9 regions of approximately-equal area.
In drawing pairs of parameters $(p_0,p_1)$, for each cell we perform rejection sampling to find a point that belongs within the cell.
% For all visualization types, we repeat the sampling process over all tasks. MSB: I think this is understood?

% MSB: defer to supplemental
% Inspired by \cite{balzer2009capacity}, we find 9 sites for the Voronoi diagram through a force-directed mechanism: starting from 9 equally spaced sites in the triangular region, in small steps, we iteratively move the site of the under-represented cell (i.e., the cell with less-than-average area) towards its over-represented neighbors, and simultaneously move the over-represented site away from its under-represented neighbors.
% By the end of each iteration, we further force the 9 sites to be symmetric along the diagonal $y = -x + 1$, by keeping the three non-coupled Voronoi sites on the diagonal and averaging the coupled sites that are not on the diagonal.

\subsection{Study Procedures}
%% participants, edu level / lay person / no assumption
%% (high level) intro / pass qual test / complete 60 examples / post - questionnaire, TLX
%% Compensation:  ~10 USD / hour ~ XYZ% of participants finished the task within (~50min, check)
%% Our human study is approved by Institutional Review Board (IRB)
%% Mixed-factor design vis type - between (three values), alignment within 
%% duplicate examples in all three types

\subsubsection{Experimental setup and recruitment procedure}

We conducted a 3 (visualization type, between) $\times$ 3 (alignment type, within) mixed-factors design for our study. We treat visualization as a between-subjects factor to ensure consistent responses, and prevent comparative judgements across visualizations. Participants complete a total of 60 trials, where we sample saliency maps of 3 different alignment types in equal 20/20/20 amounts. A single image is replicated \emph{across} visualizations, in order to permit comparison of the same image set.
To reduce confusion and repetition, each participant is assigned to at most one version of the randomly cropped images.
We recruited 90 participants in total, 49 of which recruited via Prolific \footnote{\url{www.prolific.co}}, and the remaining 41 being college students.
We limited participation in Prolific to those at least age 22, having completed at least an undergraduate degree. Though there exists some variability between these 2 populations, in our analyses we did not find significant differences between their responses. 
We assigned each participant to a unique visualization type, giving an equal 30/30/30 split between participants. Our choice of population base, e.g. laypersons with potentially limited AI background, is consistent with prior studies~\cite{10.1007/978-3-031-19775-8_17,nguyen2021effectiveness}, namely, model explanations are mainly designed to help users of diverse backgrounds solve tasks in collaboration with AI models. This study has been approved by the Institutional Review Board (IRB) at the authors' institution.

We provide instructions and qualification test before the main study -- please see supplemental material for additional details.
In the instructions, we describe the task that participants are expected to complete, and the three criteria that should be used in making judgements (see Sec.~\ref{sect:task}).
We provide participants with a qualification test across 4 trials that we determine to be saliency maps of little ambiguity -- participants must successfully complete the qualification to take part in the study.
After completing the main study (60 trials), participants are given a post-study questionnaire and a self-assessment sheet of NASA Task Load Index (TLX) \footnote{\url{https://humansystems.arc.nasa.gov/groups/tlx/}} .
Preliminary trials suggested that our full study -- instructions, qualification, and 60 trials -- required approximately one hour to complete. Commensurate with this time, we compensate participants with \$10 USD.

\section{Analyses \& Findings}

In this section we present our experimental design for analyzing the results of our study, and consequently, the findings made from our analyses. We organize the analysis by two experiments: one based on studying the design space of visualizations, while the other is focused on understanding the saliency values -- in the parameter space of visualizations -- that were chosen by participants in our study.

\subsection{Experiment 1 - Assessing the design space of saliency-based visualizations}

\begin{figure*}[tb]
    \includegraphics[width=0.9\linewidth]{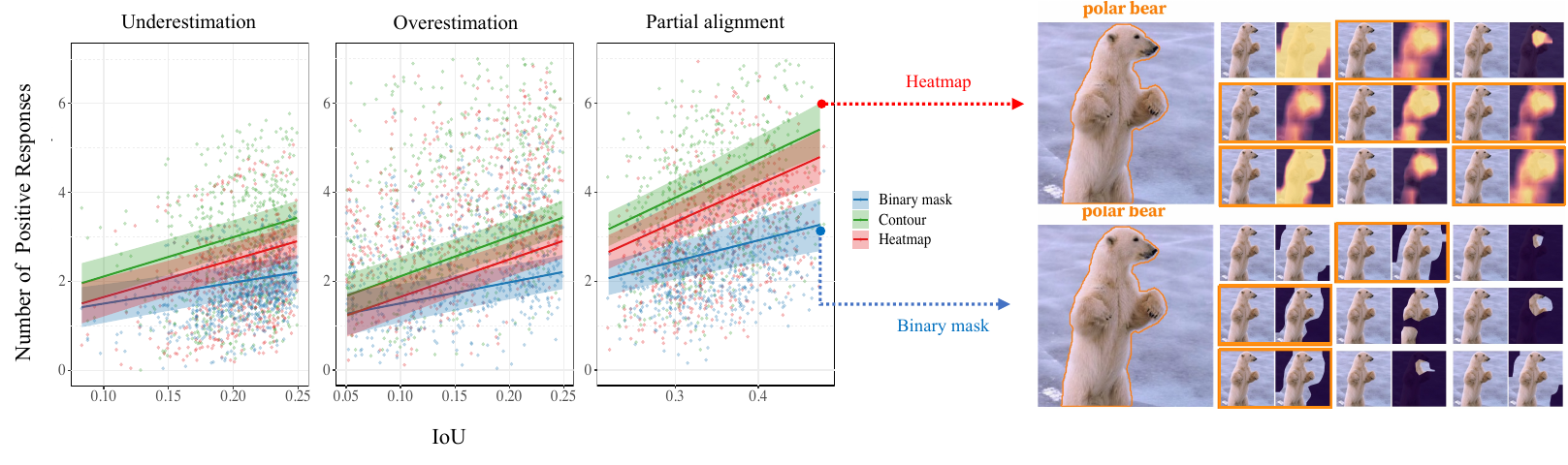}
    \caption{
    As the level of alignment (IoU) increases, we observe more positive responses for visual encodings (heatmap, contour) that show \emph{more information} about the saliency map. 
    For example, on the same saliency map, we receive 6 positive responses to heatmap visualizations (top-right), while only 3 positive responses to the binary masks (bottom-right).
    In this case, the participant of the heatmap appeared to ignore the less bright regions surrounding the "polar bear" and forgive dark regions (i.e., waist of the bear) that connect the two brighter regions (i.e., upper and lower body), whereas the participant of the binary mask considers disconnected upper and lower body in binary mask visualizations less tolerable.
    }
    \label{fig:model1-vis-alignment-iou}
    \Description[Line plots showing how number of positive responses for visual encodings of saliency maps increase with IoU increases for different alignment types.]{Left: line plots for three different alignment types (overestimation, underestimation, partial alignment) showing number of positive responses from 0 to 6 on the Y axis against IoU from 0 to 0.25 on the X axis. Among these three alignment types, we consistently observe more positive responses for visual encodings of heatmap and contour than binary mask. Right: we show examples of user study results for heatmap and binary mask visualization for the same image and saliency map.}
\end{figure*}

In this experiment, we study how the design space of a visualization -- namely, its range of possible parameter settings (c.f. Sec.\ref{sect:parameters}) -- influences participant responses. 
For a single trial, we propose to use the \emph{number of positive responses} provided by a participant as a proxy for design space coverage. The more positive responses provided, the more that a user agrees with the visualization's design space for a given saliency map.

\subsubsection{Model Design}

We employ a mixed-effects linear regression model to study what factors -- and factor combinations -- are predictive of our response variable, taken as the number of positive responses supplied by a participant in a trial. Through an iterative model expansion and refinement process \cite{gabry2017visualization}, we arrived at the following predictive model, expressed in Wilkinson-Pinheiro-Bates notation~\cite{pinheiro2017package}:
\begin{align}
\begin{split} % split environment numbers the equation only once
num\_pos =&\; vis\_type * iou \\
&+ entropy * alignment\_type * vis\_type \\
&+ shape\_simplicity \\
&+ (1+iou|user\_id)
\end{split}
\end{align}
For fixed effects we consider our individual factors, and further condition on interactions between level of alignment (IoU) and the visualization, as well as all interactions between entropy, alignment type (over/underestimate, partial alignment), and the visualization type. 
More expansive models demonstrated that IoU and alignment type did not have interactions. 
We treat participants ($user\_id$) as random effects, associated with random intercepts, as well as IoU-dependent random slopes, i.e. we expect perception of alignment to be participant-dependent. 
All post-hoc analyses are conducted using Tukeys Honest Significant Difference Test (HSD, $\alpha$ = .05).

\subsubsection{Results}

% IoU<->vis type
In comparing different visualization types, we find a significant interaction with IoU ($F(2,662) = 7, p = 0.0009$), but visualization type on its own does not have a significant effect ($F(2,332) = 1.44, p < 0.3$). 
As previously discussed (Sec. \ref{sect:categorize-alignment}), we sample saliency maps that have deliberately small IoU values, and thus, we anticipate little variation in response across visualizations when the alignment is poor. 
We expect, and observe, visualization-specific intercepts to be small, to accommodate this case. 
As the level of alignment increases, however, we observe visualization-specific slopes (c.f. Fig.\ref{fig:model1-vis-alignment-iou}) help distinguish contour ($\beta = 8.53 \pm 1.66796$) and heatmap ($\beta = 7.96 \pm 1.66796$) from binary mask ($\beta = 4.36 \pm 1.66796$). 
We find that people give more positive responses, as the level of alignment increases, for visual encodings that show \emph{more saliency detail} (\textbf{\underline{H1}}) -- please see Fig.\ref{fig:model1-vis-alignment-iou}. This suggests that participants utilize the additional information presented in contour and heatmap-based visualizations, e.g. discarding certain regions in the image as unimportant given their visual encoding, whereas binary masks are unable to discount such regions. 
We do not find a significant difference between contour and heatmap, however -- this corroborates prior studies that demonstrate quantization of color-encoded 2D scalar fields, under certain tasks, can lead to similar task performance~\cite{padilla2016evaluating}. Moreover, through our NASA-TLX survey, we find contour-based visualization ($\mu = 5.6 \pm 0.468$) require more mental effort compared to heatmaps ($\mu = 4.97 \pm 0.468$), also confirming findings in~\cite{padilla2016evaluating} regarding time required to complete tasks with quantized visual encodings.% Although not significant, we did find binary masks ($\mu = 4.70 \pm 0.468$) required the least mental effort compared to contour and heatmap visualizations, likely due to the smallest amount of information

\begin{figure*}[tb]
    \includegraphics[width=0.9\linewidth]{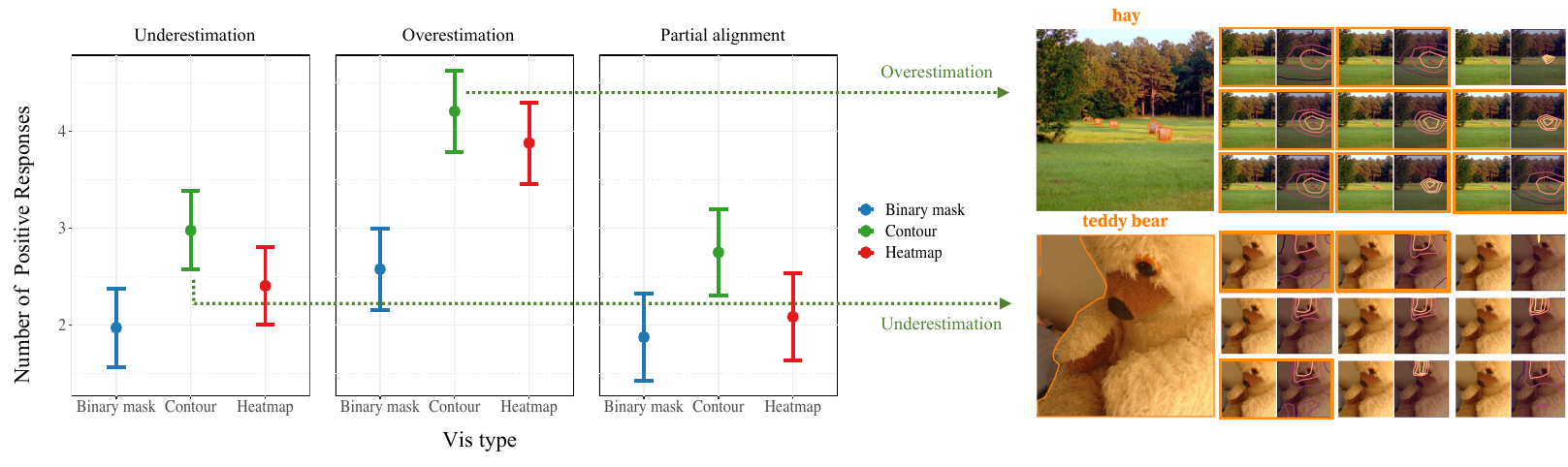}
    \caption{Across all alignment relationships, overestimated saliency maps tend to receive more positive responses than the underestimated or partially aligned instances. 
    This suggests that participants are more forgiving of imperfect alignment, so long as the object in question is covered by the saliency map visualization. 
    For example, a participant gave 8 positive responses to an overestimated saliency map (``hay'') in the contour visualization, while a separate participant provided 3 positive responses to visualizations of an underestimated saliency map (``teddy bear''). 
    In cases where the ground-truth objects are small (e.g. ``hay'' example), participants tend to find the saliency map aligns well with the ground-truth object so long as the bright contours fully enclose the ground-truth in contour visualizations.
    The preference to overestimate is more prominent for quantitatively-encoded visualizations (heatmap and contour-based visualizations) than binary masks, suggesting that participants are utilizing the additional visual cues provided by quantitatively-encoded visualizations.
    }
    \label{fig:model1-vis-alignment}
    \Description[Interval plots showing the number of positive responses for visual encodings of saliency maps for different visual encodings of saliency maps and alignment types.]{Left: Interval plots for three different alignment types (overestimation, underestimation, partial alignment) showing number of positive responses from 0 to 6 on the Y axis against three different visualization types (heatmap, binary mask, and contour) on the X axis. Across three alignment types, overestimated saliency maps tend to receive more positive responses than underestimated or partially aligned instances. Right: we show examples of user study results for contour visualization of both overestimated saliency maps and underestimated saliency maps.}
\end{figure*}  

\begin{figure*}[tb]
    \includegraphics[width=0.9\linewidth]{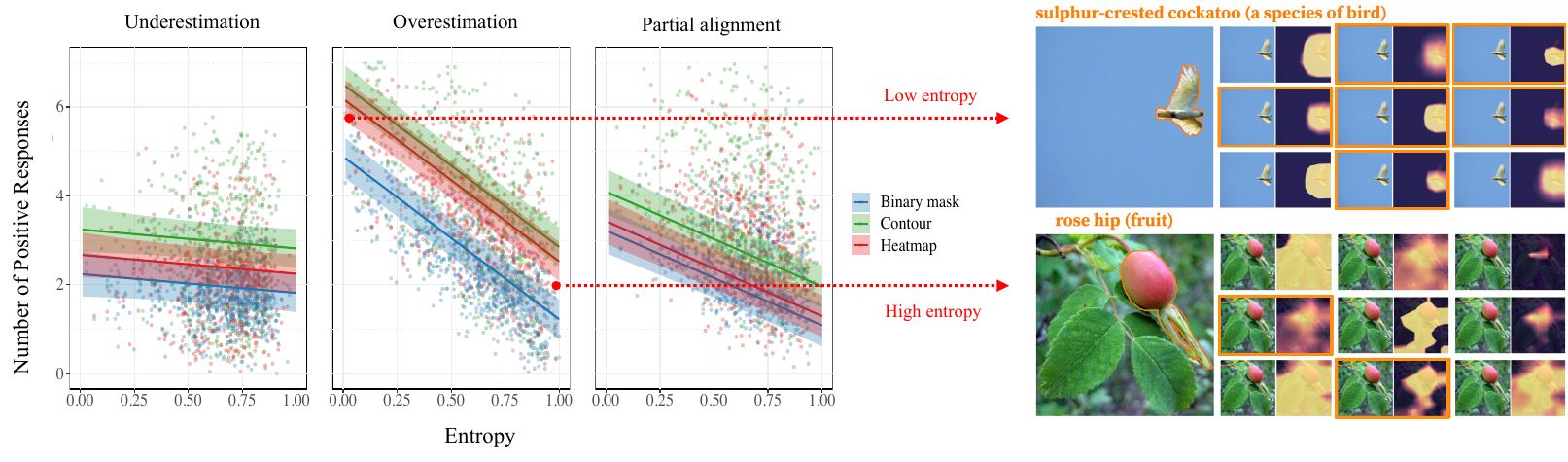}
    \caption{We observe that entropy is negatively correlated with the number of positive responses, especially for overestimated saliency maps. This is anticipated, as high-entropy saliency maps give rise to a more diverse set of visualizations in comparison to low entropy ones, and thus less number of candidates that align well with the object. For example, the saliency map overestimating "rose hip" on the image only receives 2 positive responses in its heatmap visualizations, due to its complex spatial distribution of salient regions, inclusive of the object itself. In contrast, the low-entropy saliency map overestimating the “bird” on the image receives 6 positive responses, for its visualizations in different parameters have similar appearance and all cover the "bird" on the image.
    }
    \label{fig:model1-vis-alignment-entropy}
    \Description[Line plots showing how the number of positive responses for visual encodings of saliency maps descreases as the entropy of saliency map increases.]{Left: Line plots for three different alignment types (overestimation, underestimation, partial alignment) showing number of positive responses from 0 to 6 on the Y axis against entropy of saliency map from 0 to 1 on the X axis. Across three alignment types, entropy is negatively correlated with the number of positive responses, especially for overestimated saliency maps. Right: we show examples of user study results for heatmap visualization of overestimated saliency maps with both low and high entropy.}
\end{figure*}

% alignment type, and alignment type<->vis type
We find alignment type to have a significant impact on positive responses obtained from participants \textbf{\underline{(H2)}}. We find that it is significant across its different levels ($F(2,5283) = 168, p < 0.0001$), while demonstrating significant interactions between visualization types ($F(4,5283) = 5, p < 0.0003$, c.f. Fig.\ref{fig:model1-vis-alignment}). We consistently observe that saliency maps which overestimate objects lead to a larger number of positive responses ($\mu = 3.59 \pm 0.237$), relative to underestimated saliency ($\mu = 2.41 \pm 0.229$) -- please see Fig.\ref{fig:model1-vis-alignment}. This suggests that users are more forgiving of imperfect alignment, so long as the object in question is covered by the saliency map conveyed in the visualization. The effect is most prominent for heatmap ($\mu = 3.90 \pm 0.4116$) and contour-based visualization ($\mu = 4.24 \pm 0.4116$), compared to binary mask visualization ($\mu = 2.63 \pm 0.4116$), which further supports that participants are utilizing visual cues in quantitatively-encoded visualizations, relative to binary masks.

% entropy<->vis type, entropy<->alignment type
Contrary to our initial hypothesis on entropy (\textbf{\underline{H3}}), we do not find significant interactions between entropy and visualization type. 
However, we do find that entropy and alignment type contain significant interactions (c.f. Fig.\ref{fig:model1-vis-alignment-entropy}). We observe that entropy is negatively correlated with the number of positive responses ($\beta = -0.479 \pm 0.801$); this is anticipated, as high entropy will give rise to a more diverse set of visualizations in comparison to low entropy, and thus a narrower set of consistent candidates that align with the object, demonstrated in Fig.\ref{fig:model1-vis-alignment-entropy}. Interestingly, this effect is most pronounced for overestimated saliency maps -- we find that high-entropy saliency maps that overestimate the object will give rise to a complex spatial distribution of salient regions, inclusive of the object itself. It is in these cases that participants tend to disagree with alignment, in comparison to overestimated alignments that are simpler in appearance (low entropy).

% shape
Last, we do not find an effect of shape (\textbf{\underline{H4}}) on factors of visualization type or alignment type. Interestingly, treated as an individual factor, we find that as the shape complexity of an object increases, the number of positive responses increases.

% We observe that entropy is negatively correlated with the number of positive responses ($\beta = -0.479 \pm 0.801$); this is anticipated, as high entropy will give rise to a more diverse set of visualizations in comparison to low entropy, and thus a narrower set of consistent candidates that align with the object. Interestingly, this effect is most pronounced for overestimated saliency maps -- we find that high-entropy saliency maps that overestimate the object will give rise to a complex spatial distribution of salient regions, inclusive of the object itself. It is in these cases that participants tend to disagree with alignment, in comparison to overestimated alignments that are simpler in appearance (low entropy).

\begin{figure}[!t]

    %% btw the legend here is incorrect, fixed in new version
    % \includegraphics[width=0.95\linewidth]{_vis-params-w-large.pdf}
    
    %% TODO title (Binary Mask Visualization, Contour Visualization)
    \centering
    % \includegraphics[width=0.8\linewidth]{plot_legend.png}
    % \includegraphics[trim={2cm 1cm 5cm 2cm},clip,width=0.8\linewidth]{plot_bm.png}
    % \includegraphics[trim={2cm 1cm 5cm 2cm},clip,width=0.8\linewidth]{plot_ct.png}
    % trim={<left> <lower> <right> <upper>}
    \includegraphics[trim={0 0 0cm 0.1cm},clip,width=\linewidth]{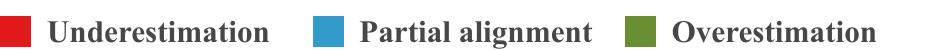}
    \includegraphics[trim={0.4cm 0.8cm 1.8cm 0.8cm},clip,width=0.8\linewidth]{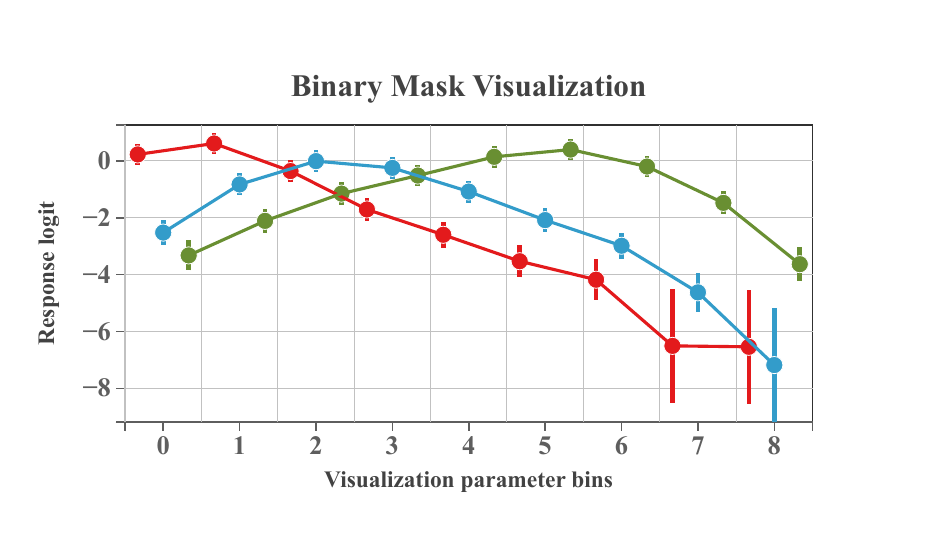}
    \includegraphics[trim={0.4cm 0.8cm 1.8cm 0.8cm},clip,width=0.8\linewidth]{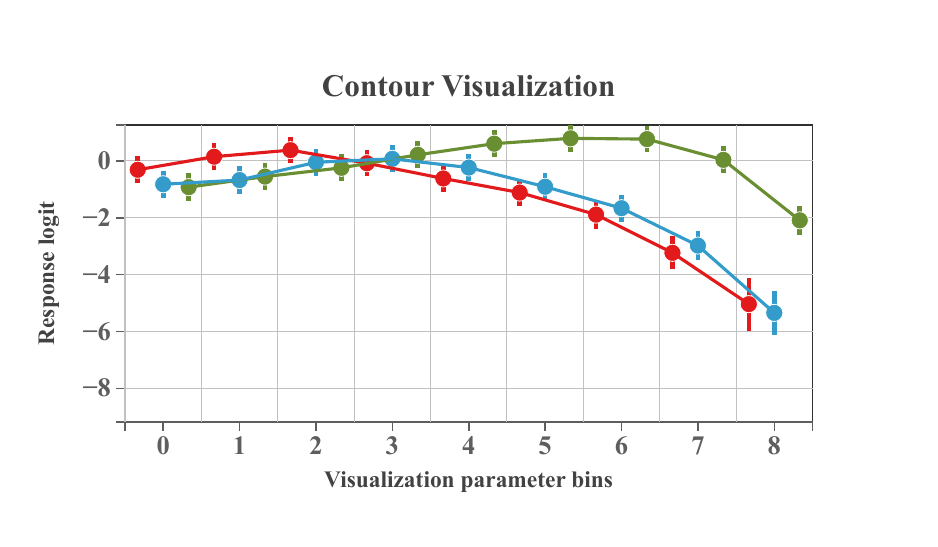}
    \caption{We show alignment-specific intercepts from our logistic regression model, for contour visualizations and binary mask visualizations. We observe consistent trends in effects of alignment type for both, while observing that humans agree on  larger parameter values in contour-based visualizations, compared to binary masks.}
    \label{fig:model2-params}
    \Description[Line plots]{Line plots showing the alignment-specific intercepts from our logistic regression model for binary mask visualization and contour visualization. The Y axis showing the alignment-specific intercepts ranging from -8 to 0. The X axis showing the visualization parameter bins from 0 to 8. Three lines for three different alignment types are shown in each graph. We consistently observe that humans agree on larger parameter values across three alignment types in contour-based visualizations, compared to binary mask visualization.}
\end{figure}

\subsection{Experiment 2 - relating human judgement with visualization designs}

In our second experiment we seek a more specific connection with human judgements made on saliency alignment, and the design parameters that underlie the visualizations chosen by participants. Our goal is to improve the understanding of visualization parameters that are predictive of individual participant responses, conditioned on factors that describe the level of alignment (IoU), saliency map (entropy), and alignment type. To this end, we propose to build \emph{visualization-specific} logistic regression models that take as input such factors, in addition to a \emph{parameter bin} (c.f. Sec.\ref{sect:parameters-sample}) associated with the visualization. For instance, given that our visualization design presents 9 visualizations to participants in a single trial, a single bin for a binary mask would comprise the range of values between $[0,\frac{1}{9}]$ from which thresholds for the mask were sampled. The purpose of the model is to best represent binary judgements provided by participants in our study.

\subsubsection{Model Design}

An immediate challenge that we must handle arises from our interface design: we anticipate that human judgement on selecting one visualization depends on the remaining visualizations in the presented $3 \times 3$ grid. Thus, models that assume independence amongst observations are inappropriate. To this end, we use the model presented in~\cite{brooks2017glmmtmb} for performing mixed-effects logistic regression, which permits the direct specification of covariance structures across a subset of input variables. We supply a Gaussian kernel for measuring proximity of binned value ranges in the parameter space, e.g. adjacent bins will have a high similarity, bins far apart will have a similarity that decays to zero. Overall, our model can be expressed as follows, again following Wilkinson-Pinheiro-Bates notation~\cite{pinheiro2017package}:
\begin{align}
\begin{split} % split environment numbers the equation only once
logit(x) =&\; iou + entropy \\
&+ p\_bin * alignment\_type \\
&+ (0+\Sigma(p\_bin)|user\_id)
\end{split}
\end{align}
where $x$ contains all predictors, and we use the notation $\Sigma(p\_bin)$ to indicate a prescribed, inter-participant covariance matrix prior for the binned parameter ($p\_bin$) indicator variables. 
A sigmoid transformation is applied to this response, for maximization of a (regularized) Bernoulli likelihood~\cite{brooks2017glmmtmb}. 
We choose to study interactions only between alignment types, and binned parameters, as we expect our binning is sufficiently fine-grained that factor-specific slopes across binned parameters would offer little contribution.

\subsubsection{Results}

We first study models for contour-based visualizations and binary mask visualizations, as they share the same parameter space. 
For both binary mask visualization 
% ($\beta = 0.52179 \pm 0.11156, p < 0.0001$) % (old) link function = gaussian
($\beta = 3.9872 \pm 0.4402, p < 0.0001$) % (new) link function = binomial
and contour visualizations 
% ($\beta = 0.98973 \pm 0.1259, p < 0.0001$),
($\beta = 6.0596 \pm 0.3981, p < 0.0001$), 
we find significant positive effects of IoU on a human's positive response; 
we also find a significant negative effect of entropy of saliency map on a human's positive response when visualized with a binary mask 
% ($\beta = -0.2742 \pm 0.02658, p < 0.0001$) 
($\beta = -2.0529 \pm 0.1052, p < 0.0001$) 
or contour 
% ($\beta = -0.30724 \pm 0.03, p < 0.0001$).
($\beta = -1.8463 \pm 0.0954, p < 0.0001$).
We find significant effects of visualization parameters for contour 
% ($\chi^2(8) = 149, p < 0.0001 $) 
($\chi^2(8) = 179.90, p < 0.0001 $) 
and binary mask 
% ($\chi^2(8) = 324, p < 0.0001$), 
($\chi^2(8) = 164.32, p < 0.0001$), 
in addition to significant interactions between parameters and alignment types, for contour 
% ($\chi^2(16) = 1033, p < 0.0001$) 
($\chi^2(16) = 896.07, p < 0.0001$) 
and binary mask 
% ($\chi^2(16) = 3054, p < 0.0001$). 
($\chi^2(16) = 1748.737, p < 0.0001$). 

\begin{figure}[!t]
    \includegraphics[width=0.8\linewidth]{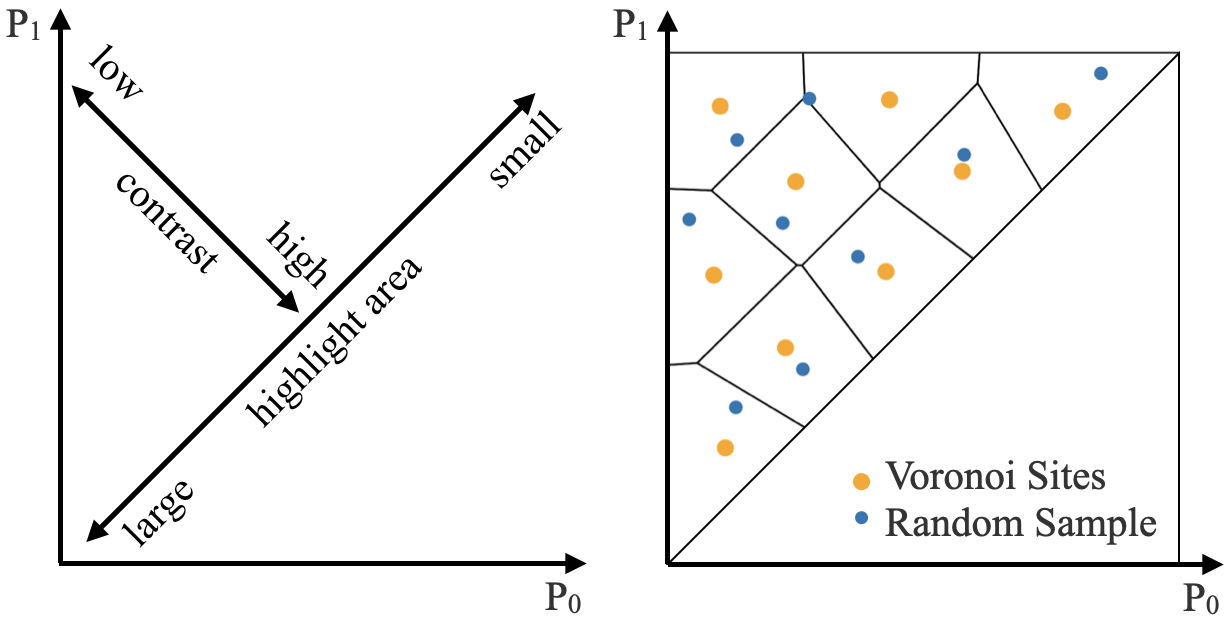}
    \includegraphics[width=\linewidth]{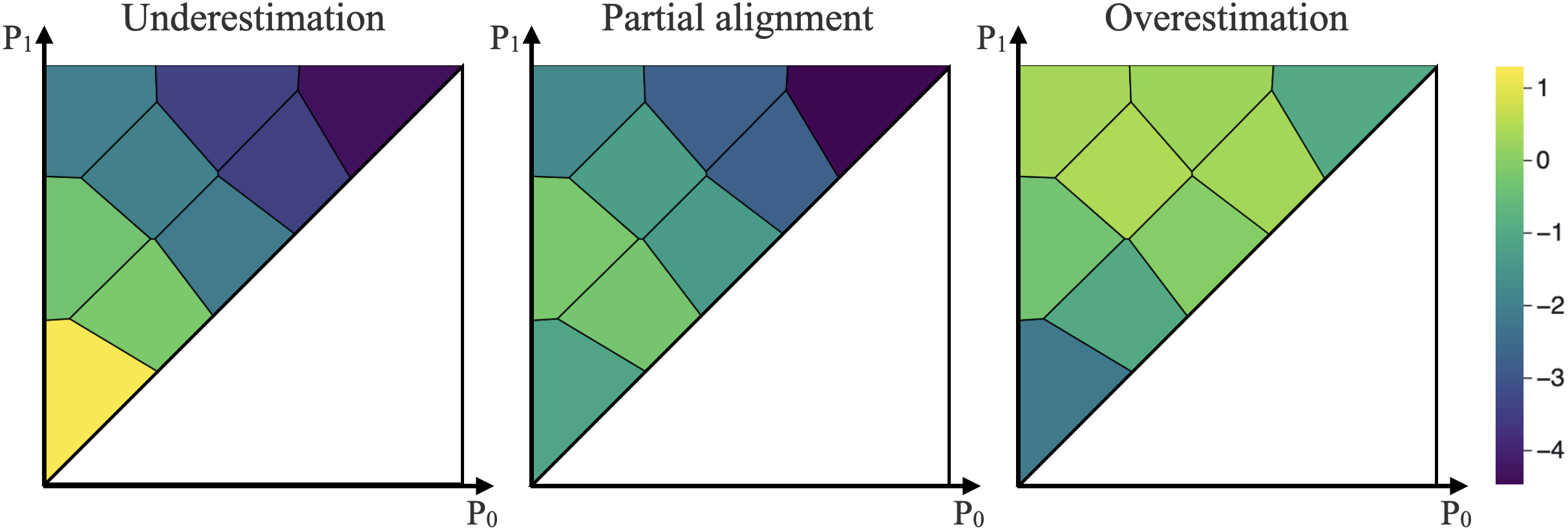}
    
    \caption{
        % \textbf{Left to right:}
        \textbf{Top:}
        Visual effects and the centroidal Voronoi tessellation on the heatmap design space.
        \textbf{Bottom:}
        Results from experiment 2: alignment-specific intercepts as color-coded cells within clipped Voronoi diagrams.
    }
    \label{fig:model2-alignment}
    \Description[Plots for visual effects and centroidal Voronoi tessellation]{Top left: a plot showing how 1) the area of highlighted area on the visualization of saliency map 2) the contrast of visualization of saliency maps change along the axises of P0 and P1 (two data filterint parameters for visualization of saliency map). Top right: a plot showing the centroidal Voronoi tessellation on the heatmap design space. Bottom: graphs for three different alignment types showing the alignment-specific intercepts as color-coded celss within clipped Voronoi diagrams.}
\end{figure}

% In Fig.\ref{} we plot the intercepts for the parameters across all bins in the contour visualization, where bars are colored and stacked by alignment type. 

In Fig.\ref{fig:model2-params} we plot the intercepts for parameters across all bins in the binary mask and contour visualizations, where error bars are colored and linked by alignment type. 
In both cases, we find the parameter peaks are shifted by alignment type: the underestimated saliency peaks at the lowest parameter bin among the three, followed by partial alignment, while overestimated saliency peaks at larger parameters. 
This analysis calibrates the alignment type to the most preferable visualization design parameter: when the saliency map covers more than the object, and thus overestimates, a larger design parameter naturally shrinks the saliency view.
% Similarly, in Fig.\ref{} we plot intercept parameters for the binary mask.
In comparing the different visualizations, we see for binary masks the rate of decay peaks earlier than the contour, indicating that contour-based visualizations permit the setting of larger parameters for which humans agree.

\begin{figure}[tb]
    % \includegraphics[width=1.0\linewidth]{_application-interface-22.pdf}
    % \includegraphics[width=1.0\linewidth]{_application-interface-23.png}
    % \captionsetup[subfigure]{labelformat=empty}
    % trim={<left> <lower> <right> <upper>}
    \begin{subfigure}{\linewidth}
    \centering
        \includegraphics[trim={0 2cm 0 2.5cm},clip,width=0.9\linewidth]{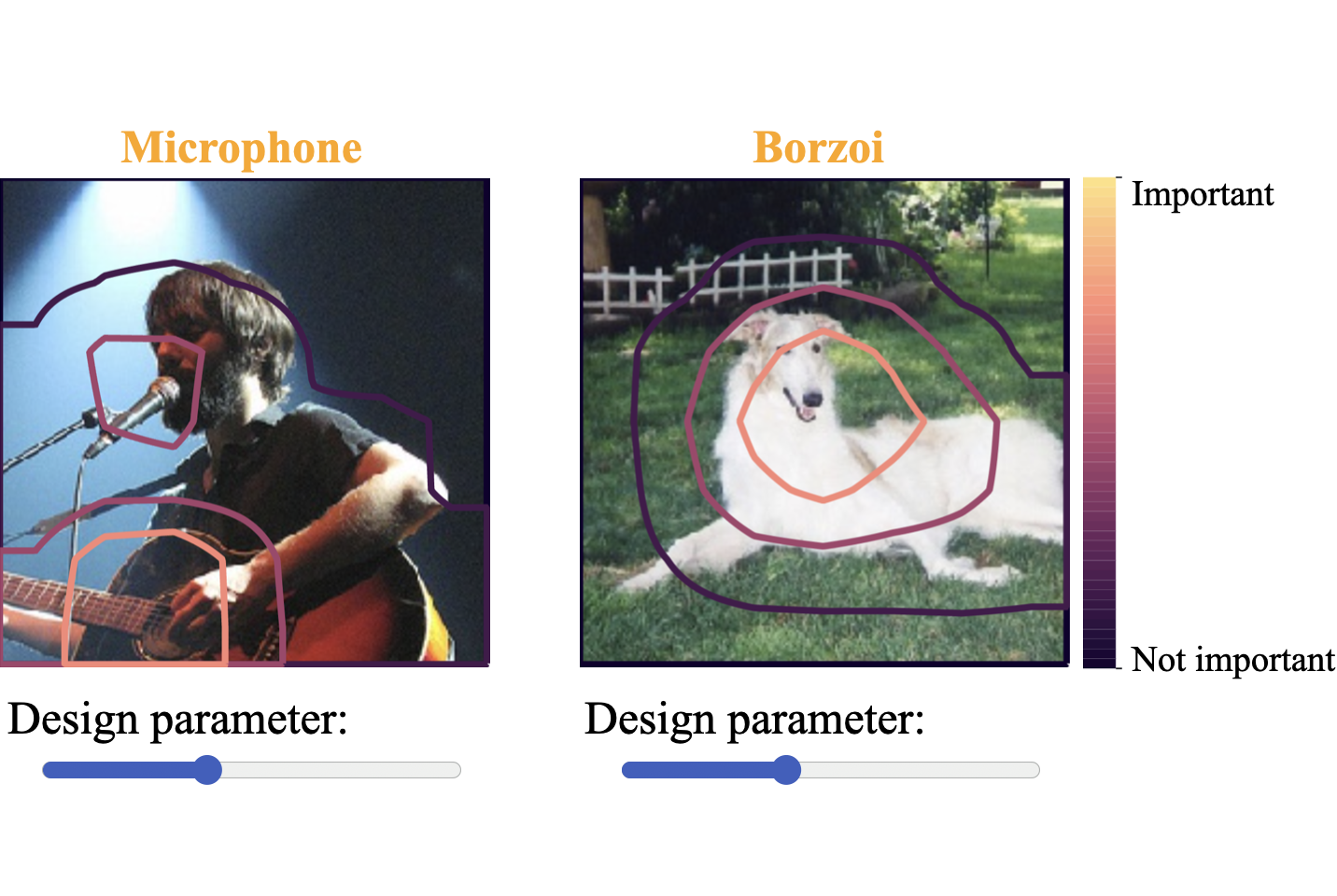}
        \caption{Conventional design}
    \end{subfigure}
    \begin{subfigure}{\linewidth}
        \centering
        \includegraphics[trim={0 5cm 0 1cm},clip,width=0.9\linewidth]{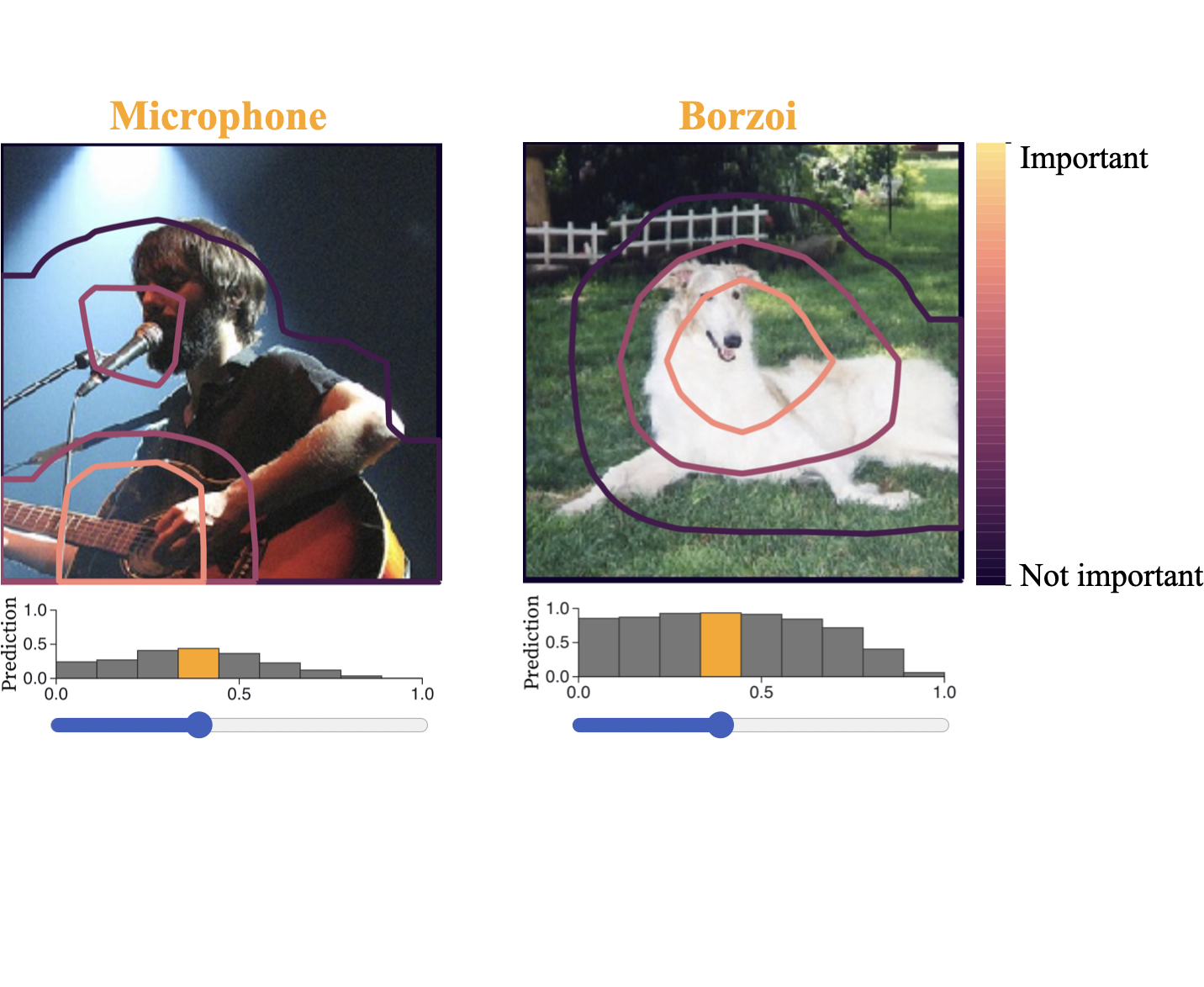}
        \caption{Informed visualization design}
    \end{subfigure}
    \caption{
    Top: conventional visualization design leaves designers uninformed about users response to specific design. 
    Bottom: designers could adjust the visualization parameter (e.g. minimum saliency value to draw contours for contour visualization) across the whole design space through the slider under the image. The predicted human response under the selected visualization design would be highlighted in the barplot. Our human perception model can make the design process more informed by showing predictions of users' responses under different designs.
    }\label{fig:application-interface}
    \Description[Conventional and our visualization design interface for saliency maps.]{Top: two examples of conventional visualization design interface. For each example, we show a) image; 2) label for the object on the image; 3) visualization of saliency map for the given object, overlayed on the image; 4) a design parameter slider below the image; 5) a color legend for the visualization of saliency map. Bottom: two examples of our visualization design interface, with expected human response shown as histogram for each design parameter choice.}
\end{figure}

The heatmap visualization presents analysis challenges due to its 2-dimensional parameter space. 
To this end, we visually encode alignment-specific intercepts as color-mapped cells within a clipped Voronoi diagrams, see Fig.\ref{fig:model2-alignment}. 
Each cell comprises a binning of the parameter space from which parameters $(p_0,p_1)$ are drawn. 
As shown in the figure, and verified in our analysis, we find significant effect of visualization parameters 
% ($\chi^2(8) = 229.870, p < 0.0001$), 
($\chi^2(8) = 90.47, p < 0.0001$), 
and their interaction with alignment type 
% ($\chi^2(2) = 681, p < 0.0001$). 
($\chi^2(2) = 1643.83, p < 0.0001$). 
As shown in the figure, we find a, generally, increasing trend of intercepts as we traverse underestimated, partially-aligned, and overestimated saliency maps. This indicates that the parameter representing the minimum saliency to clamp ($p_0$) contributes most strongly to human judgement. 
Moreover, the left-diagonal of each diagram corresponds to regions where the parameters $(p_0,p_1)$ strictly increase in their difference, moving from the center to the upper-left. 
Restricted to this region, we find that participants are \emph{consistent in their judgement}, independent of the data range for heatmap visualization and alignment type. 
This suggests showing the whole range of values (e.g. $[0,1]$), as is convention in visual explanations~\cite{selvaraju2017grad,Fel2021WhatIC}, can lead to similar responses as parameter settings whose average values are approximately equal (e.g. 0.5).

% Make it a new section?
\subsubsection{Interactive visualization prototype}
% To demonstrate the benefits of our perceptual model, 
% To demonstrate \new{prospective} benefits of our perceptual model \new{in the future}, 
To demonstrate the potential use case of our perceptual model, 
we have built an interactive prototype intended for visualization designers to explore the design space of visualizations in communicating saliency maps, please see Fig.\ref{fig:application-interface}. 
Given an input image, its ground-truth object label, and its saliency-based explanation, our prototype shows the predictions of human responses under the binned parameter space for a given visualization type. 
We augment a standard slider-based interface, one that permits a user to set the parameters of a visualization, with ``scents''~\cite{willett2007scented} -- presented as a bar plot -- corresponding to our model's prediction on whether a specific parameter setting would result in human agreement of alignment. 
This helps \emph{contextualize} visualization parameters, such that visualization designers can anticipate how their design choices would evoke responses from a target audience.

As an illustrative example in Fig.\ref{fig:application-interface}(b, left), for the image with ``microphone'' as ground-truth object, our perceptual model predicts that humans tend to think the given saliency map does not faithfully align with the ground-truth object ``microphone'' in the image, no matter how we adjust the visualization parameter for the contour-based visualization. 
As shown, the saliency map captures both the hand and the microphone, indicating a poor alignment to the target concept. 
On the other hand, as shown in Fig.\ref{fig:application-interface} (b, right), our model indicates that a wide range of parameter settings would evoke positive responses from humans. 
In either case, we emphasize that our perceptual model does not indicate precisely \emph{what} parameter should be set, nor a notion of optimal parameter configuration. 
% Value judgements of saliency explanations ultimately reside with the designer, wherein our model, and interactive interface, provides additional context to support users in making design choices.
Value judgements of saliency explanations ultimately reside with the designer. Our model, when incorporated into a designer's interface, can provide additional context to support users in making design choices.

\section{Discussion}

In this work, we studied how visualization design influences human perception of model explanations in the form of saliency maps.  In particular, our work focused on saliency maps that are likely to provoke uncertainty in human judgement, specifically, when a user is tasked
to make an assessment on how well a saliency map aligns with an object in an image. Through our user studies, we found factors pertaining to (1) visualization design decisions, (2) the alignment relationship between saliency map and the object, (3) the distribution of values in the saliency map, and (4) interactions thereof, all play an important role in how humans make judgements on saliency alignment.

Our results challenge the convention that automatic alignment scores~\cite{DBLP:conf/cvpr/BauZKO017,fong2018net2vec} are sufficient measures to be used as part of evaluating model explanations~\cite{10.1007/978-3-031-19775-8_17,nguyen2021effectiveness}. Indeed, as shown in Fig.~\ref{fig:model1-vis-alignment-iou}, considerable variation in human responses exist within a narrow range of IoU scores. These results underscore the importance of graphical perception when evaluating model explanations, in order to obtain a better understanding of human performance in AI-assisted tasks. In turn, a natural application of our study is to employ our perceptual models of alignment in place of automatic alignment scores. This has the potential to more precisely discern \emph{why} certain visual explanations result in poor human performance, e.g. a lack of alignment, rather than a misleading explanation.

For visualization designers, our findings can inspire the future visualization design of saliency explanations for different scenarios. 
Our human perception model, and interactive interface prototype, can further be used to design more human-understandable saliency explanations for model behavior in machine teaching applications, such as teaching crowd workers to complete image annotation tasks using the visualization of saliency maps as interpretable explanations~\cite{pmlr-v84-chen18g,mac2018teaching}.
We leave it to future work to further develop the interface and evaluate its effectiveness through user studies.

Beyond supporting visualization designers who aim to communicate visual explanations (c.f. Fig.~\ref{fig:application-interface}), ML researchers may also leverage our results as part of experimental design in evaluating explanations. We have shown that visualization design of saliency maps is an integral piece to understanding how humans analyze and complete AI-assisted tasks. Future human-centered evaluation studies can incorporate our perceptually-grounded visualization design into their procedure, in order to compensate for different kinds of visualizations, e.g. parameter settings for binary masks, versus contours or heatmaps.

The results of our study can further support ML researchers in the design of explanation methods. Analogous to Fel et al.~\cite{fel2022harmonizing}, the results of our work can be used to score the alignment of visual explanations in a manner that is consistent with human perception. Recent work~\cite{rigotti2021attention} has investigated the incorporation of such ``side information'' to make explanation methods more interpretable; our perceptual models can further ensure that the design of explanation methods are informed by how humans perceive visual explanations.

% a paragraph on the limitations of our study
Under the scope of this work, our study is limited to Grad-CAM~\cite{selvaraju2017grad} as saliency based explanations which produces a low-resolution saliency map. 
We believe our general methodology could be applied to other saliency methods \cite{9010039,ancona2017towards,petsiuk2018rise}, particularly those of higher spatial resolution. Furthermore, our user study is limited to participants with little or unknown AI background, instead of ML researchers. ML researchers with rich AI background, on the other hand, might have a different perception model of saliency explanations, leveraging prior knowledge on model behavior.  During our study, we also found certain human responses that were hard to explain. There might be some other factors, such as categories of objects, that our current factors cannot explain for those potential effects on humans. Lastly, we have limited measures, other than a qualification test before the main study, to make sure that participants fully understand the task through our instructions. In the worst-case scenario, human responses we collect might be noisy or exist ambiguity, though in practice the potential for such noise did not impact the statistical significance of our results.

% a paragraph on future directions
For the future improvement of our study, we intend to explore human perception of the visualization of other types of explanations for visual recognition models, or a broader range of machine learning models \cite{10.1145/3491102.3501965,10.1609/aaai.v33i01.33016309} where the visualization of model explanations plays an important role in communicating with humans. We also plan to integrate perceptual studies grounded in visualization as part of  human centered evaluation of explanations, as well as consider aspects of human perception in the design of explanation algorithms or interpretable models~\cite{rigotti2021attention}. Last, we plan to further explore visualization designs that incorporate perceptual models of saliency-based explanations, and study their effects on how designers make decisions in visually encoding saliency maps.

\section{Conclusions}

% a paragraph on conclusions

% here's what we did
% we study the factors, and do user study, we build predictive model (dependent variable, the meaning ) to understand 
In this work, we studied how visualization design impacts humans' graphical perception of saliency maps, which encodes the explanations of visual recognition models. 
In particular, we focused on saliency maps that are likely to provoke uncertainty, and studied their visual encodings along two axes: the choice of the encoding’s visual range, such as the visualization type; and the choice of its data domain for visualization - such as using visualization parameter to select an interval of saliency values to be displayed.
Our findings show that factors related to visualization design decisions, the type of alignment, and qualities of the saliency map all play important roles in how humans perceive saliency-based visual explanations.
The results of our study have implications on how visualization designers make decisions on visually encoding saliency-based explanations, alongside implications on ML researchers for designing human-centered evaluation of model explanations, wherein our findings and perceptual models permit a more precise evaluation, and understanding, of human performance in AI-assisted tasks.

%\new{Based on these factors, we build a perceptual model of human judgement on the saliency map, and also an interactive prototype for visualization designers of saliency explanations. Our findings and perceptual model can inform visualization design of saliency explanations for machine teaching~\cite{mac2018teaching}, and also inspire the design of more human-aligned explanations or interpretable model. The results of our study also present a more nuanced view on humans' decision making process in AI-assisted scenarios, as well as the human evaluation of visual explanations -- First, human perception of saliency maps cannot merely be reduced to automatic evaluation metrics, many other factors like visualization design also impact human perception; Second, thuman-centered evaluation of explanations could incorporate perception study grounded in visualization into their procedure, and decompose their evaluation results into multiple phases.} 

\bibliographystyle{ACM-Reference-Format}
\bibliography{bib}

% \appendix
  
% \input{samples/appendix}

\end{document}